\documentclass[lettersize,journal]{IEEEtran}
\usepackage{amsmath,amsfonts}
\usepackage{algorithmic}
\usepackage{algorithm}
\usepackage{array}
\usepackage[caption=false,font=footnotesize,labelfont=rm,textfont=rm]{subfig}
\usepackage{textcomp}
\usepackage{stfloats}
\usepackage{url}
\usepackage{verbatim}
\usepackage{graphicx}
\usepackage{pdfpages}
\usepackage{cite}
\begin{document}

\title{A Real-time Faint Space Debris Detector With Learning-based LCM}

\author{Zherui Lu, Gangyi Wang, Xinguo Wei, and Jian Li
\thanks{This work was supported in part by the National Key Research and Development Program of China under Grant 2019YFA0706002. \textit{(Corresponding author: Gangyi Wang.)}}
}

\markboth{Journal of \LaTeX\ Class Files,~Vol.~14, No.~8, August~2021}%
{Shell \MakeLowercase{\textit{et al.}}: A Sample Article Using IEEEtran.cls for IEEE Journals}

\IEEEpubid{0000--0000/00\$00.00~\copyright~2021 IEEE}

\maketitle

\begin{abstract}
With the development of aerospace technology, the increasing population of space debris has posed a great threat to the safety of spacecraft. However, the low intensity of reflected light and high angular velocity of small debris impede the extraction. Besides, due to the limitations of ground observation methods, small space debris can hardly be detected, making it necessary to enhance the spacecraft’s capacity for space situational awareness (SSA). Considering that traditional methods have some defects in low-SNR target detection, such as low effectiveness and large time consumption, this paper proposes a method for low-SNR streak extraction based on local contrast and maximum likelihood estimation (MLE), which can detect spatial objects with SNR 2.0 efficiently. In the proposed algorithm, local contrast will be applied for crude classifications, which will return connected components as preliminary results, then MLE will be performed to reconstruct the connected components of targets via orientated growth and the precision can be further improved. The algorithm has been verified with both simulated streaks and real star tracker images, and the average centroid error of the proposed algorithm is close to the state-of-the-art method like the ODCC. At the same time, the algorithm in this paper has significant advantages in efficiency compared with the ODCC. In conclusion, the algorithm in this paper is of high speed and precision, which guarantees its promising applications in the extraction of high dynamic targets. 
\end{abstract}

\begin{IEEEkeywords}
high dynamic streaks, local contrast map, maximum likelihood estimation, onboard camera, space object extraction.
\end{IEEEkeywords}

\section{Introduction}
\IEEEPARstart{T}{he} number of resident space objects (RSO) in orbit, especially all kinds of space debris, is constantly increasing\cite{johnson2010orbital, flury2000searching}. In 2013, there were more than 20,000 space debris with a diameter larger than 10cm identified and surveilled by NASA, of which only about 1,000 were artificial satellites\cite{kennewell2013overview}. Uncooperative space objects have posed a huge challenge to the security of spacecraft in orbit. Due to the limitations of ground observation, space debris with a diameter of less than 10cm is difficult to identify and track, and space debris of this size is enough to be a fatal menace for spacecraft\cite{schildknecht2007optical, silha2017optical}. Therefore, space-based detection methods have become a new direction in space research, and the space situational awareness (SSA) capability of spacecraft has also become a popular research topic at present\cite{weeden2010global, munir2022situational, lal2018global}. One of the feasible methods is to detect based on optical devices such as onboard cameras, which are of low cost and high efficiency. Besides, as a mature platform, there will be less risk and difficulties in the development and implementation of new algorithms. Thus, space target detection based on the optical method is one of the most widely used detection methods currently\cite{sharma2000space}. 
As the space debris is small, the intensity of reflected light can be weak\cite{schildknecht2003optical}, which conduces to its low signal-to-noise ratio (SNR). Moreover, space debris usually has a large angular velocity and presents high dynamic characteristics in the image plane, which further reduces the SNR\cite{zhang2012blurred, ma2016region, sun2013motion}. In addition, the hardware resources of the in-orbit platform are limited, constraining the algorithm design and deployment. These above factors make it more challenging to detect high-dynamic and low-SNR targets in orbit. 
At present, there are many common methods of low-SNR object extraction, which are mainly based on the gray distribution difference between the object and the background or the gray distribution characteristics of the object itself to achieve object extraction. The former relies on the discontinuity of gray distribution patterns lying in the target and its neighborhood, and contrast is performed for image enhancement. In 2012, Chengqiang Gao developed an extraction method for small targets via sparse ring representation, which can achieve stronger image enhancement and a higher detection rate compared to other methods\cite{gao2012small}. Philip Chen raised a new method known as local contrast map (LCM) in 2013, and the method realized image enhancements and noise suppression at the same time by calculating the difference between the target region and the background region. At the end of the procedure, an adaptive threshold will be adopted for image segmentation\cite{chen2013local}. Chen’s algorithm involved less calculation and reached high extraction precision and is one of the typical algorithms of target extractions. Then in 2018, Jinhui Han proposed an object extraction method based on relatively local measurements. In his method, multi-scale templates were introduced, which presented higher robustness and a more efficient computation process for spatial objects with different sizes\cite{han2018infrared}. However, the methods above concentrate on extractions of small targets, and the capability of extractions is constrained by the templates involved in some of these methods. When it comes to extended targets, such as streaks, there will be a considerable degradation in their capabilities.
\IEEEpubidadjcol
The latter takes advantage of the gray characteristics of the target itself, such as the shape and direction constraints, the probability density function (PDF) of the gray distribution, and so on, to extract targets. For instance, in 2015, Fujimoto designed an extraction method for dim spatial targets based on the track-before-detection (TBD) method, which can effectively extract and track spatial targets subject to specific distribution by combining multiple frames\cite{fujimoto2015statistical}. However, such methods require prior knowledge of the target energy distribution function and much computation. In 2017, according to the property that maximum curvature exists at the endpoints of the streak, Sease proposed an endpoint extraction method based on second-order gradient features to determine the range of the streak\cite{sease2017automatic}. However, this detection method is not applicable under the condition of low SNR, because the morphological characteristics of the streak are very insignificant. The deblurring method for star points under high dynamic conditions based on fuzzy kernel was proposed by Jiang Jie in 2016\cite{jiang2016accelerated}, which is more efficient than the traditional method of star point restoration, but it has higher requirements on computation and is consequently less efficient compared to other extraction methods. One of the classical methods in the field of high dynamic object extraction is to enhance the image via a convolution template, which has been proven to be the optimal detection method. In 2014, Hou Wang proposed an enhancement method based on directional integration\cite{hou2014smeared}, that is, integrating along the direction of star movement to improve the SNR of the target. However, the integration direction of this method is fixed, which is inconsistent with real situations, and it is only applicable to simple cases. To further improve the quality of the connected component, the ODCC method proposed by Xiaowei Wan in 2019\cite{wan2021odcc}, added MLE to achieve the detection of ultra-low SNR spatial targets, of which the lowest detectable SNR is 1.4 and the average centroid error is less than 1 pixel.

ODCC method can effectively identify spatial targets with different SNRs and achieve target segmentation. However, while looking into its process, this method involves multi-directional convolution filtering and a non-maximum suppression process for each time of filtering. In addition, this method carries out a traversal search for regions with possible targets to obtain the global optimal solution. This process can be time-consuming and requires a lot of hardware resources to run. Additionally, the start point of the traversal search is largely determined by the mixed Gaussian model, which may not be accurate at low SNR and the search results may be adversely affected. 

Given the above shortcomings of the ODCC methods, this paper proposes a fast extraction method of extremely low-SNR space objects based on the idea of local contrast and the orientated growth of connected components. The algorithm first extracts gray features within a template, then trains an SVC to classify the central pixel of the template and obtains a crude connected component. Finally, by utilizing MLE, orientated growth of the target connected component is performed to determine the end points of the streak and reconstruct the complete connected component.

The paper is organized as follows: the first part is the background introduction of feasible methods, the second part is the imaging analysis of stars under high dynamic conditions, the third part is the illustration of the proposed method in detail, the fourth part is the parameter selection and condition setting of the simulation and experiment, the fifth part is the test results with the simulation and real star map, the sixth part is the conclusion, and the last part is the further discussion on the proposed algorithm.

\section{Characterization of High Dynamic Streaks}
\subsection{Energy Distribution of Streak}
For satellite-borne cameras, the imaging characteristics of RSOs far away from the camera are very similar to real stars, so the energy distribution analysis for star points is also applicable to RSOs. A defocusing design is adopted to reach higher precision of measurement\cite{liebe1998active}, thereby the energy distribution of a star spot at static approximately follows a two-dimensional Gaussian distribution. However, when under dynamic conditions, especially high dynamic, the star spot will pass through multiple pixels within the exposure time\cite{wei2014exposure}. As a result, a streak will be presented on the imaging plane, rather than a single spot. As the energy of a star is dispersed into pixels that the star has passed by, the SNR of a dynamic star can be lower than which at static. The energy distribution of the streak can be expressed as
\begin{equation}
    I_s(x_i, y_i) = \frac{I_c}{2\pi\sigma^2}\int_0^T{\exp{\frac{(x_i-x_c(t))^2+(y_i-y_c(t))^2}{2\sigma^2}}}dt
\end{equation}
where $I_c$ and $\sigma$ are the parameters of the descriptive function, $T$ denotes the exposure time of the camera and $(x_c(t), y_c(t))$ represents the energy centroid at time $t$. The signal-to-noise ratio (SNR) of a star is usually expressed as the peak signal-to-noise ratio (PSNR)\cite{barniv1987dynamic}, which is defined by
\begin{equation}
    \mathrm{PSNR}=\frac{g_\mathrm{max}-g_\mathrm{bck}}{\sigma}
\end{equation}
where $g_\mathrm{max}$ represents the gray of the brightest pixel in the target region, $g_\mathrm{bck}$ refers to the average gray of the background and $\sigma$ denotes the standard deviation of the background noise.
\subsection{Onboard Camera Under High Dynamic Conditions}
\begin{figure}[ht!]
    \centering
    \subfloat[]{
        \includegraphics[width=0.45\columnwidth]{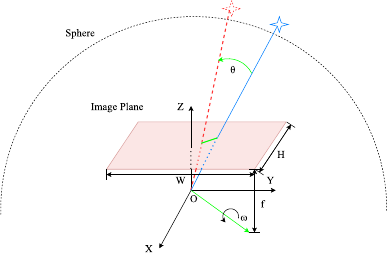}}
    \subfloat[]{
        \includegraphics[width=0.45\columnwidth]{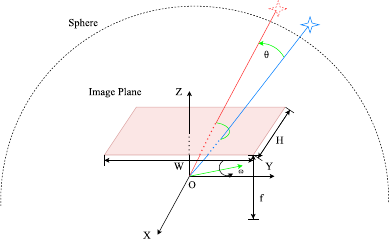}}
    \caption{Imaging model of star sensors under dynamic conditions. (a) The angular velocity of the star sensor has no component on the Z-axis. (b) The angular velocity of the star sensor has a component on the Z-axis.}
    \label{fig:1}
\end{figure}

The trace left by the spatial object on the image plane is determined by the relative motion between the object and the onboard camera, as shown in Fig.~\ref{fig:1},
\\where $\omega$ denotes the angular velocity vector of the onboard camera, $W$ and $H$ represent the width and height of the image plane respectively, $f$ is the focal length and $\theta$ denotes the angle at which the star is rotated relative to the center of light at time $t$.

\begin{figure}
    \centering
    \includegraphics[width=0.45\columnwidth]{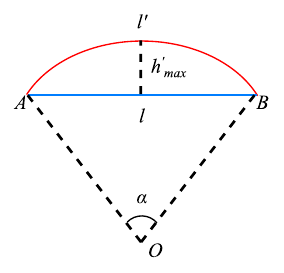}
    \caption{The relationship between $l^{'}$ and $l$. $l^{'}$ denotes the arc $AB$, $l$ denotes the chord $AB$, points $A$ and $B$ represent the endpoints of the chord and $\alpha$ represents the rotated angle of the star within the exposure time.}
    \label{fig:2}
\end{figure}

As illustrated, the trajectory of the object on the image plane is straight when the angular velocity of the camera has no component on the Z-axis. However, when the star sensor rotates around the Z-axis, the trajectory on the image plane can be regarded as a conic curve. According to the conclusion of Xiaowei Wan, the difference between the star point trajectory and the straight-line segment can be evaluated by the arch height $h_\mathrm{max}^{'}$, as shown in Fig.~\ref{fig:2}.
The arc $AB$ can be considered as a straight line approximately when $h_\mathrm{max}^{'}$ is small enough, and the constraint can be deduced subsequently as follows:

\begin{equation}
    \alpha\leq2\arccos{(1-\frac{\cos^4{\Phi}}{f/h_\mathrm{max}^{'}})}
\end{equation}
where $\Phi$ denotes the half field of vision (FoV), $f$ denotes the focal length of the camera and $h_\mathrm{max}^{'}$ is set to 1.

\section{Methodolgy}

\begin{figure}
    \centering
    \includegraphics[width=0.8\columnwidth]{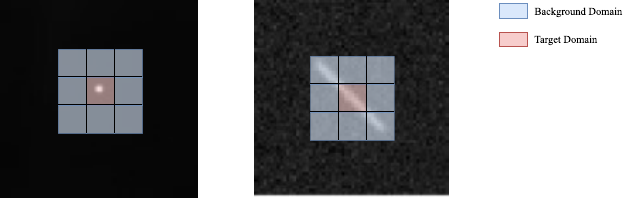}
    \caption{Traditional LCM template with spot and streak targets.}
    \label{fig:3}
\end{figure}

LCM is a typical method of image enhancement, as introduced in Part I. This algorithm adopts the average gray of the target neighborhood $m_i(i=1,2,\dots,8)$ and the brightest gray of pixels within the target sub-region $L_n$ as features to describe the discontinuity of gray distribution. The local contrast map can be obtained by

\begin{equation}
    C_n=\min_i{L_n\times C^n_i}=\frac{L^2_n}{m_i}
\end{equation}
where $C_i^n=\frac{L_n}{m_i}$ denotes the gray contrast of the $i$th sub-region and the obtained $C_n$ will replace the central pixel of the target region. The process of LCM is very simple and thus of good efficiency. Besides, the algorithm can achieve image enhancement and noise suppression simultaneously via
$C_n$. However, the extraction capacity of LCM will degrade significantly when extended targets like streaks are encountered, as shown in Fig.~\ref{fig:3} since such targets will pose an influence on target neighborhoods and thus reduce the $C_n$.

\begin{figure}
    \centering
    \includegraphics[width=1\columnwidth]{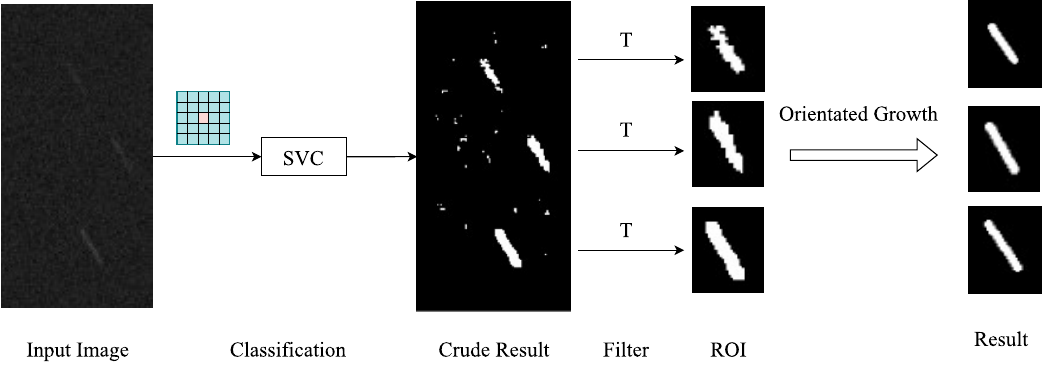}
    \caption{Process of the proposed algorithm}
    \label{fig:4}
\end{figure}

Although the streak excesses the target area, not all sub-regions are affected, as shown in Fig.3. For example, the sub-regions in the lower left and upper right corners can be considered as pure background areas so that local contrast is still feasible for enhancement and extraction theoretically. The more pure background areas in the template, the less influence caused by the extended target on the template region, and the better the effect of local contrast will be. Therefore, a bigger template has more advantages in the capacity of detection than a smaller template. Additionally, as the uninfluenced sub-regions are not fixed due to the various directions and lengths of the streaks, it is unrealistic to distinguish the pure background sub-regions for each case. With the analysis above, a support vector machine (SVM) is introduced for automatic selection, and then local contrast and orientated growth will be performed to obtain complete connected components of targets. The process of the algorithm is shown in Fig.~\ref{fig:4}.

The algorithm consists of two stages, the first is crude classification and the last is orientated growth. The former extracts the gray features from the template and an SVC is trained to classify the central pixel of templates. Then criteria like the size of connected components are adopted to get rid of misclassification caused by noise. The latter performs orientated growth on connected components that are believed to be targeted via the morphology constraints of the streak. Finally, the connected components of targets are reconstructed completely.
\subsection{Crude Classification Based on Local Contrast}
\begin{figure}
    \centering
    \includegraphics[width = 0.5\columnwidth]{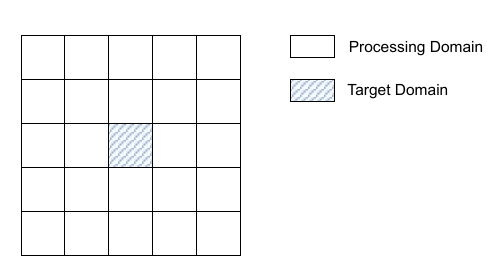}
    \caption{The template used in the proposed algorithm.}
    \label{fig:5}
\end{figure}

The proposed algorithm adopts a $25\times25$ template, which can be divided evenly into $25$ sub-regions with a scale of $5\times5$, as shown in Fig.~\ref{fig:5}.

At first, the average gray of each sub-region $m_i(i=1,2,\dots,25)$ and the brightest gray of the target region $g_\mathrm{max}$ will be extracted as features. So, the feature space $X$ of the proposed algorithm has 26 dimensions in total.
\begin{equation}
    X=\{x_1, x_2, \dots, x_{26}\}
\end{equation}
To improve the robustness towards different background patterns, the minimum average gray feature will be utilized as an evaluation of local background $m_\mathrm{bck}$, which will be subtracted from other features.
\begin{gather}
    m_\mathrm{bck}=\min_i m_i(i=1,2,\dots, 25)\\
    x_i=m_i-m_\mathrm{bck}(i=1,2,\dots, 25)\\
    x_{26}=g_\mathrm{max}-m_\mathrm{bck}
\end{gather}

Then an SVC with a Gaussian kernel is trained to classify the central pixel in the template, and the process can be expressed by
\begin{equation}
    f(x)=\boldsymbol{\omega}^T\phi(\boldsymbol{x})+b
\end{equation}
\begin{equation}
    y=
    \begin{cases}
        1 & f(\boldsymbol{x})\geq T_h \\
        0 & f(\boldsymbol{x})\leq T_h
    \end{cases}
\end{equation}
where $\boldsymbol{x}$ denotes the feature vector extracted from the template, $\phi(\boldsymbol{x})$ represents the feature vector derived from the mapping of $\boldsymbol{x}$, $T_h$ denotes the threshold of the classifier. Theoretically, the central pixel will be more likely to be classified as the target if the streak has passed through the target region of the template, which is responsible for the widening of the connected components in the crude classification stage. Such an effect is identical to dilation to some extent, reducing the potential of fracture occurrence and guaranteeing the robustness of the proposed algorithm at extremely low SNR.

\begin{figure}
    \centering
    \includegraphics[width=\columnwidth]{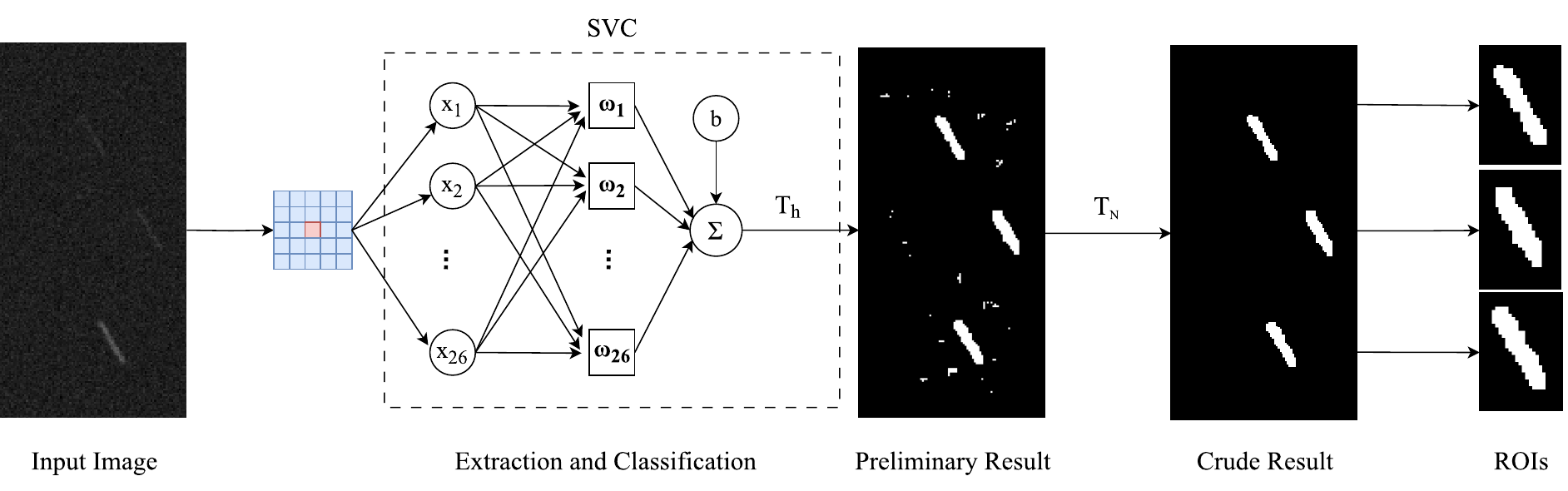}
    \caption{Framework of crude classification. A template is adopted for feature extraction and an SVC is trained for central pixel classification.}
    \label{fig:6}
\end{figure}

A binary map will be obtained after the traversal of the whole image and a series of connected components that consist of both background and target pixels are extracted. Generally speaking, the connected component formed by noise pixels is usually more scattered and contains far fewer pixels than the real target. Therefore, a simple criterion can be set to mark connected components that are large enough as potential targets, and the process of the crude classification is shown in Fig.~\ref{fig:6}.
Above all, the process in detail can be summarized as Algorithm ~\ref{alg:1}.
\begin{algorithm}
    \caption{Crude Classification Process}
    \label{alg:1}
    \begin{algorithmic}
        \STATE 
        \STATE {\textbf{Input: } raw image $R$}
        \STATE {\textbf{Output: } rough results set $T_r$}
        \STATE {\text{Create template $M$, empty sets $T_r=\{\}$, $T=\{\}$}}
        \WHILE{not traversed $R$}
            \STATE {$m_i=\frac{1}{25}\sum_j^{25} g_j^i$\quad\text{for} $i=1,2,\dots,25$}
            \STATE {$m_\mathrm{bck}=\min_i m_i (i=1,2,\dots,25)$}
            \STATE {$x_i=m_i-m_\mathrm{bck}$\quad\text{for} $i=1,2,\dots,25$}
            \STATE {$\boldsymbol{x}=\{x_1,x_2,\dots,x_{25}, g_\mathrm{max}-m_\mathrm{bck}\}$}
            \STATE {$f(x)=\boldsymbol{\omega}^T\phi(\boldsymbol{x})+b$}
            \IF{$f(\boldsymbol{x})\geq T_h$}
            \STATE {$y=1$}
            \ELSE
            \STATE {$y=0$}
            \ENDIF
        \ENDWHILE
        \FOR{$j=1,2\dots,\vert T\vert$}
        \STATE {\text{Extract connected components and add them to set $T$}}
        \IF{sizeof$(t_j)\geq t_h$}
        \STATE {{$T_r=T_r\cup \{t_j\}$}}
        \ENDIF
        \ENDFOR
        \STATE {\text{Return }$T_r$}
        \end{algorithmic}
\end{algorithm}

\begin{figure}
    \centering
    \subfloat[]{\includegraphics[width=0.8\columnwidth]{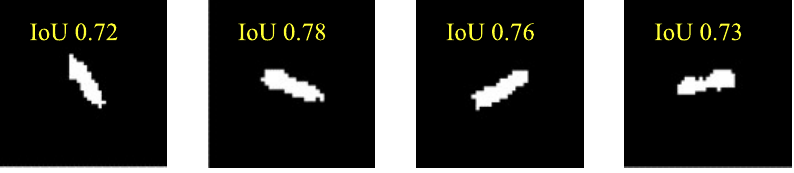}}\\
    \subfloat[]{\includegraphics[width=0.8\columnwidth]{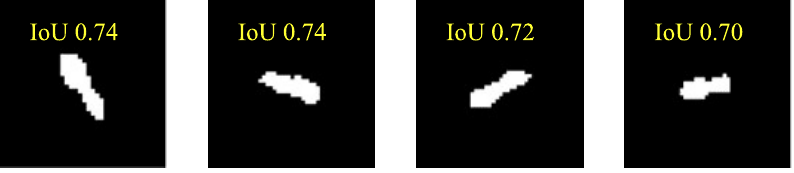}}\\
    \subfloat[]{\includegraphics[width=0.8\columnwidth]{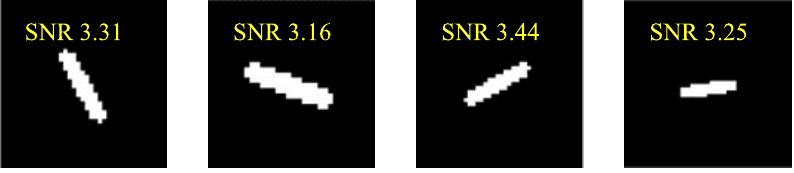}}
    \caption{ Results of the crude classification process. (a) The results of crude classification process with Gaussian kernel. (b) The results of crude classification process with linear kernel. (c) The ideal connected components of targets}
    \label{fig:7}
\end{figure}

However, the SVC with Gaussian kernel is too complicated to be implemented as the number of support vectors exceeds 2,200, and the increase in computation time is unacceptable. Therefore, an SVC with a linear kernel is adopted as a substitution and an experiment is carried out to compare the effect of SVCs with linear and Gaussian kernels, which is illustrated in Fig.~\ref{fig:7}.

\begin{figure}
    \centering
    \includegraphics[width=0.7\columnwidth]{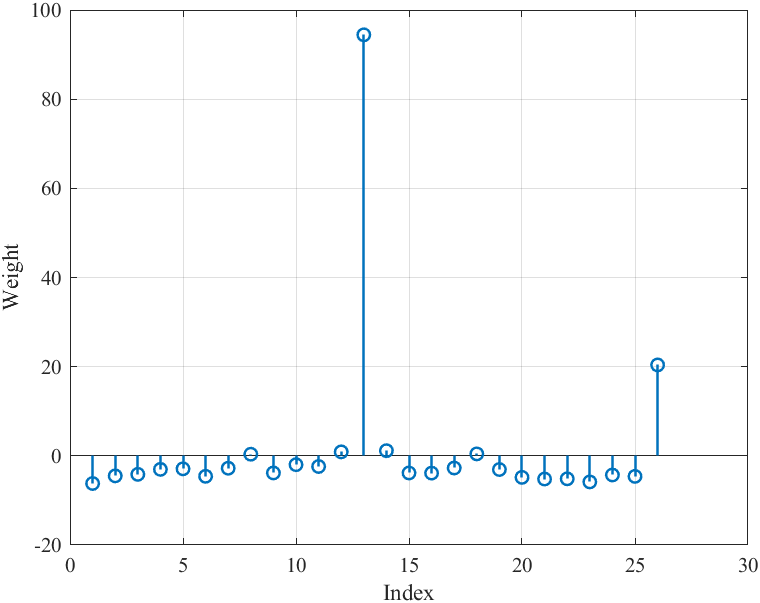}
    \caption{The weight vector of the trained SVC with linear kernel. The figure manifests that the features extracted from the central region are of the highest weights. In other words, these two features distinguish the target pixel from the background one.}
    \label{fig:8}
\end{figure}

According to Fig.7, the Intersection over Unions (IoUs) of two different kernels are close, so these two methods can be equivalent in this case. The SVC with a linear kernel involves far less calculation and requires fewer hardware resources, which makes it more efficient and feasible than that with a Gaussian kernel. The weight of each feature in SVC with a linear kernel is shown in Fig.~\ref{fig:8}.

\begin{figure}
    \centering
    \includegraphics[width=0.6\columnwidth]{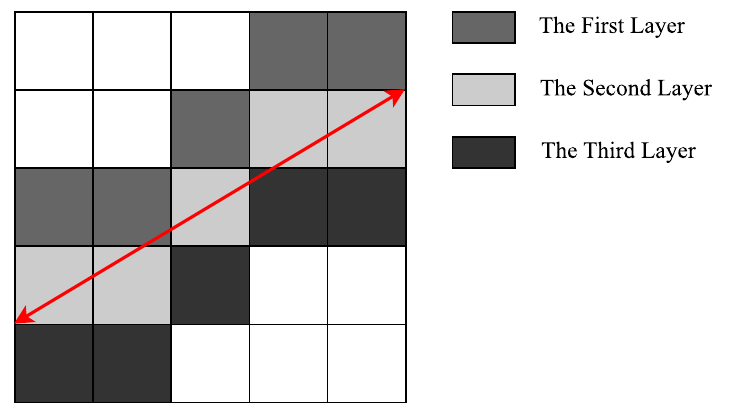}
    \caption{ The distribution of layers in the target sub-region.}
    \label{fig:10}
\end{figure}

With the presumption that the background noise $g_N\sim N(\mu_N, \sigma_N^2)$ and pixels in the template are mutually independent, the capability of the proposed template can be evaluated by theoretical analysis. For simplicity, the width of the streak is assumed to be 3 pixels. According to the mixed Gaussian model, the 3-layer model is established to describe the streak, noted as $l_1$, $l_2$, and $l_3$, respectively. Then the PDF of each layer can be expressed by
\begin{equation}
    f(g_{ij};\mu_i, \sigma_i)=\frac{1}{\sqrt{2\pi}\sigma_i}\exp{\frac{(g_{ij}-\mu_i)^2}{2\sigma_i^2}}
\end{equation}
where $g_\mathrm{ij}\in l_i(i=1,2,3)$, $\sigma_i$ denotes the standard deviation of the $i$th layer, $\mu_i$ denotes the mean gray of the $i$th layer and $g_\mathrm{ij}$ denotes the gray of the $j$th pixel in the $i$th layer. Consequently, the average gray $\Bar{g}_S$ of a sub-region $S$ can be represented by
\begin{equation}
    \Bar{g}_S=\frac{1}{N_0}(\sum_\mathrm{g \in BG}g+\sum_\mathrm{g\in S\cap l_1}g+\sum_\mathrm{g\in S\cap l_2}g+\sum_\mathrm{g\in S\cap l_3}g)
\end{equation}
where $N_0=25$ represents the number of pixels that a sub-region contains, and $BG$ denotes the set that consists of background pixels in $S$. The distribution that a background sub-region $S_b$ follows and its parameters are expressed as follows:
\begin{gather}
    \mu_{S_b}=\mu_N\\
    \sigma_{S_b}^2=\frac{\sigma_N^2}{N_0}\\
    \Bar{g}_{S_b}\sim N(\mu_{S_b}, \sigma_{S_b}^2)
\end{gather}
Similarly, the distribution that the target sub-region follows and its parameters are expressed as follows:
\begin{gather}
    \mu_{S_t}=\frac{1}{N_0}(\vert BG \vert\mu_N+\sum_{i=1}^3 \vert S_t\cap l_i \vert \mu_i)\\
    \sigma_{S_t}^2=\frac{1}{N_0^2}(\vert BG \vert\sigma_N^2+\sum_{i=1}^3 \vert S_t\cap l_i \vert \sigma_i^2)\\
    \Bar{g}_{S_t}\sim N(\mu_{S_t}, \sigma_{S_t}^2)
\end{gather}
For the highest gray $g_m$ of the target sub-region, its possibility distribution function $F_\mathrm{max}(g_m)$ can be represented by
\begin{align}
    F_\mathrm{max}(g_m) & = \prod_\mathrm{i=1}^{25} P\{g_i \leq g_m\}\notag\\
    & = \prod_\mathrm{g_m\in BG}\Phi(\frac{g_m-\mu_N}{\sigma_N})\cdot\prod_\mathrm{i=1}^3\prod_\mathrm{g_m\in l_i}\Phi(\frac{g_m-\mu_i}{\sigma_i})
\end{align}
and the corresponding PDF can be represented by
\begin{equation}
    f_\mathrm{max}(g_m)=\frac{dF_\mathrm{max}(g_m)}{d(g_m)}
\end{equation}
Let the first $k$ elements of the features vector be the mean grays of the background sub-regions, then the weighted sum $a$ of each feature can be obtained by
\begin{align}
    a & = \sum_\mathrm{i=1}^{26} \omega_i x_i + b\\
    & = \sum_\mathrm{i=1}^k\omega_i\Bar{g}_{S_{bi}}+\sum_\mathrm{i=k+1}^{25}\omega_i\Bar{g}_{S_{ti}} + \omega_{26}x_{26}+b\\
    & = A_1 + A_2 + b
\end{align}
where 
\begin{equation}
A_1=\sum_{i=1}^k\omega_i\Bar{g}_{S_{bi}}+\sum_{i=k+1}^{25}\omega_i\Bar{g}_{S_{ti}}
\end{equation}
\begin{equation}
    A_2=\omega_{26}x_{26}
\end{equation}
The term $A_1$ follows Gaussian distribution apparently and its expectation and standard deviation are calculated by
\begin{gather}
    \mu_{A_1}=\mu_{S_b}\sum_{i=1}^k\omega_i+\sum_{i=k+1}^{25}\omega_i\mu_{S_{ti}}\\
    \sigma_{A_1}^2=\sigma_{S_b}^2\sum_{i=1}^k\omega_i^2+\sum_{i=k+1}^{25}\omega_i^2\sigma_{S_{ti}}^2\\
    A_1\sim N(\mu_{A_1}, \sigma_{A_1}^2)
\end{gather}
However, the term $A_2$ doesn’t follow any common distributions, which makes it difficult for theoretical analysis. Therefore, the effect of the crude classification is evaluated by the PDFs of the main features instead, and further research is carried out with the streaks whose directions are set to be $30^\circ$. Each layer in the target sub-region is shown in Fig.~\ref{fig:10}.

\begin{figure}
    \centering
    \includegraphics[width=0.8\columnwidth]{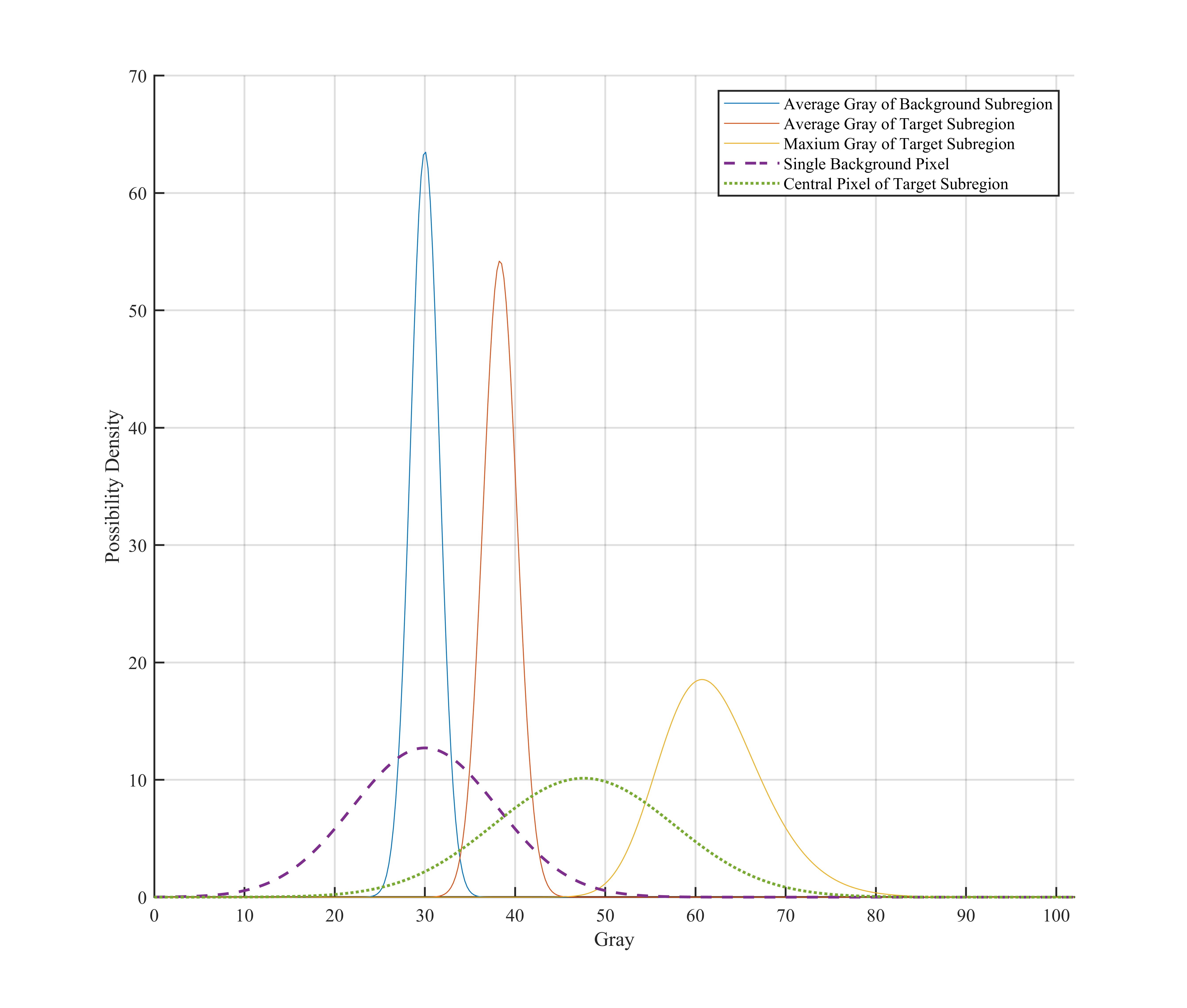}
    \caption{The PDFs of $\Bar{g}_{S_b}$, $\Bar{g}_{S_t}$ and $g_m$, compared with these of single background pixel and the central pixel in the target sub-region}
    \label{fig:11}
\end{figure}

Then the PDFs of $\Bar{g}_{S_b}$, $\Bar{g}_{S_t}$ and $g_m$ can be plotted respectively as Fig.~\ref{fig:11}, where the PDF of $g_m$ is attained by the Monte-Carlo method. 
While comparing with the background pixel and the central pixel, it turns out that the mean features are of significantly higher possibility densities, compacting the distribution into a smaller range and steeping the curves. Besides, when looking at the PDFs of the mean features, these three curves have little overlap with each other, which attests that $g_m$ and $\Bar{g}_{S_t}$ are distinguishing enough to indicate whether a target exists in the target sub-region. In contrast, the PDFs of the background pixel and the central pixel have a large common part with each other, presenting the fact that these kinds of features are not capable of classification.

\subsection{Connected Component Orientated Growth}
\begin{figure}
    \centering
    \includegraphics[width=0.9\columnwidth]{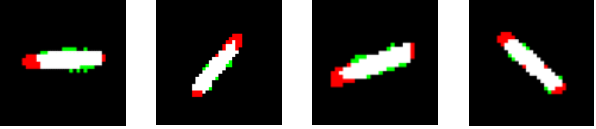}
    \caption{Comparison between the results of crude classification and ideal connected components. Streaks are generated with random lengths and directions, and red indicates the missing pixels while green refers to the redundant pixels. }
    \label{fig:12}
\end{figure}

According to Fig.~\ref{fig:7}, the edges of connected components are not smooth, and pixels that are redundant or missing will contribute to centroid errors. Even worse, the defects of connected components may fail the extractions. Background noise is responsible for such defects, as its influence becomes more significant at the target edge when under low SNR conditions. For a more intuitive understanding, an experiment is designed to compare the ideal connected component with the results returned by the crude classification, as shown in Fig.~\ref{fig:12}.

As shown in Fig.~\ref{fig:12}, the extracted results tend to have missing pixels at both ends of the streak, while the redundant pixels are concentrated on both sides of the streak. In addition, the length of the missing part is usually very short, so it is unnecessary to carry out a traversal search like the ODCC method to obtain the globally optimal result. Based on the analysis above, an orientated growth method based on MLE\cite{cooper1979maximum} is proposed, which utilizes the direction and shape constraints of streaks to reconstruct the complete connected component of the target. 

\begin{figure}
    \centering
    \includegraphics[width=0.4\columnwidth]{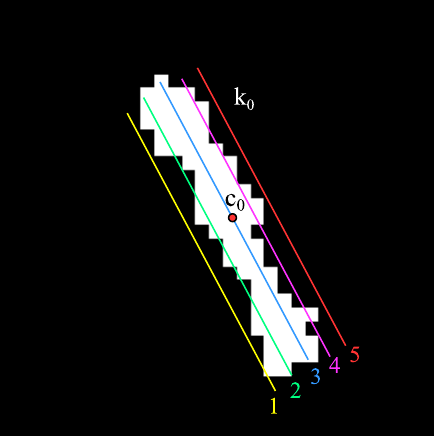}
    \caption{Mixed Gaussian model with five layers.}
    \label{fig:13}
\end{figure}

Firstly, the initial slope $k_0$ can be calculated via extractions of the centroid $c_0$ and the pixel furthest from $c_0$. Then a 5-layer mixed Gaussian model (GMM) \cite{dementhon1995model} is established with the middle layer passing through $c_0$, as shown in Fig.~\ref{fig:13}.
With the formula (11), the PDF of each layer of the Gaussian model can be expressed by
\begin{equation}
    p(g_{ij};\mu_i, \sigma_i) = \frac{1}{\sqrt{2\pi}\sigma_i}\exp{\frac{(g_{ij}-\mu_i)^2}{2\sigma_i^2}}
\end{equation}
where $i\in \{1,2,3,4,5\}$, the layer on the bottom is defined to be the first layer, while the top one is defined as the fifth layer.

\begin{figure}
    \centering
    \includegraphics[width=0.4\columnwidth]{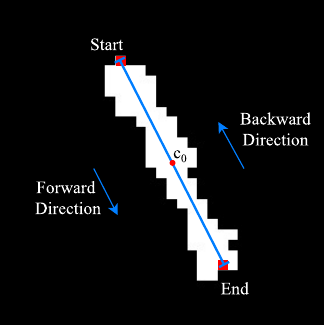}
    \caption{Definitions of the endpoints and the orientation of the streak.}
    \label{fig:14}
\end{figure}

Considering the centroid error, the central axis of the connected component is not necessarily to be the third layer. Based on the distribution characteristics of streaks, the layer with the highest mean gray is selected, say the $m$th layer, to be the central axis, then the joint probability density (JPD) of the central axis can be obtained by
\begin{equation}
    p(g_m;\mu_m, \sigma_m)=\prod_j \frac{1}{\sqrt{2\pi}\sigma_m}\exp{\frac{(g_{mj}-\mu_m)^2}{2\sigma_m^2}}
\end{equation}
where $g_{mj}$ denotes the $j$th pixel in the central axis.
Subsequently, a search for the optimal slope is proceeded, in which the neighborhood of $k_0$ is traversed by a fixed step, and calculate the JPD of each step. However, the length of the central axis can vary with the slope, so that the average joint possibility density (AJPD) is more appropriate for the estimation, whose definition is represented by
\begin{equation}
    \Bar{p}=\sqrt[n_m]{p(g_m; \mu_m, \sigma_m)}
\end{equation}
where $n_m$ denotes the number of pixels that the central axis contains. The formula (28) can be expressed in logarithmic form as follows:
\begin{equation}
    \ln{\Bar{p}}=\frac{1}{n_m}\sum_j\frac{(g_{mj}-\mu_m)^2}{2\sigma_m^2}-\ln{\sqrt{2\pi}\sigma_m}
\end{equation}
After traversal, the slope $k_{max}$ with the highest AJPD is chose as the slope of the current connected component. Taking account of streak width, the m-1th and the m+1th layers are also selected apart from the central axis, and these layers will be adopted as the seed of orientated growth. The JPD of seed can be represented by
\begin{equation}
    p_\mathrm{seed}=\prod_{i=m-1}^{m+1}\prod_{j=1}^{n_i} \frac{1}{\sqrt{2\pi}\sigma_i}\exp{\frac{(g_{ij}-\mu-I)^2}{2\sigma_i^2}}
\end{equation}
where $n_i$ denotes the number of pixels the $i$th layer contains. As pixels are mutually independent and the seed is considered as a part of the target, only the JPD of the newly grown section needs to be focused on. The definitions of streak direction and endpoints are shown in Fig.~\ref{fig:14}.

\begin{figure}
    \centering
    \includegraphics[width=\columnwidth]{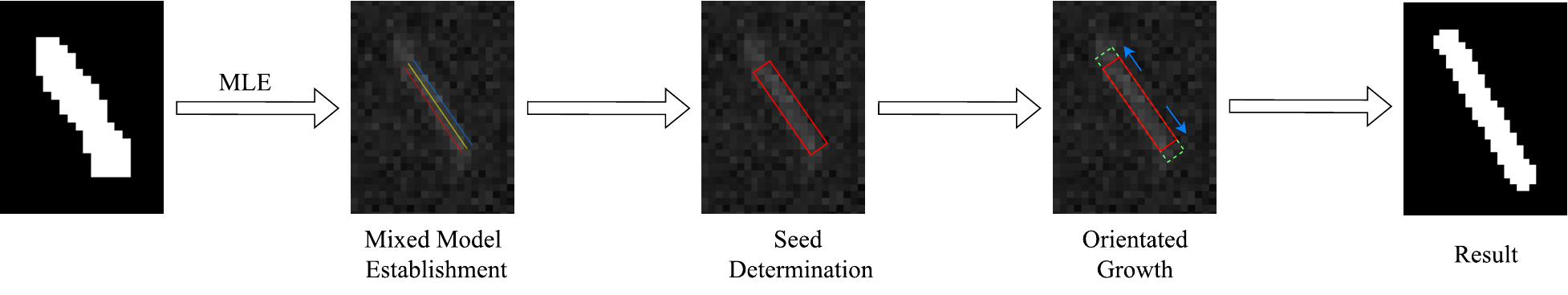}
    \caption{Process of orientated growth algorithm.}
    \label{fig:15}
\end{figure}

The connected component will grow along the forward direction at first, and one pixel will be attached to each layer in each step, say $a_i (i\in\{m-1, m, m+1\})$. Then the JPD of these newly attached pixels that all belong to target $p_\mathrm{all\_t}$ and the JPD of these newly attached pixels that all belong to background $p_\mathrm{all\_t}$ can be expressed as follows:
\begin{gather}
    p_\mathrm{all\_t} = \prod_{i=m-1}^{m+1}\frac{1}{\sqrt{2\pi}\sigma_i}\exp{\frac{(a_i-\mu_i)^2}{2\sigma_i^2}}\\
    p_\mathrm{all\_b} = \prod_{i=m-1}^{m+1}\frac{1}{\sqrt{2\pi}\sigma_b}\exp{\frac{(a_i-\mu_b)^2}{2\sigma_b^2}}
\end{gather}
If $p_\mathrm{all\_t} > p_\mathrm{all\_b}$, the newly attached pixels will be considered as a part of the target, and the seed will be extended subsequently. Otherwise, the seed will stop growth and reverse the growth direction. An up limit $L_\mathrm{max}$ is also set for maximum growth length, and the growth process will stop when the length of the grown section $L$ reaches $L_\mathrm{max}$. Thus, the stop conditions of the algorithm can be concluded as follows:
\begin{equation}
   \left\{
    \begin{array}{cc}
         p_\mathrm{all\_t} &> p_\mathrm{all\_b}, \\
         L &> L_\mathrm{max} .
    \end{array}\right.
\end{equation}
Then the connected component will grow along the backward direction, and the above process will be repeated until the stop conditions are met. The process of orientated growth is shown in Fig.~\ref{fig:15}.

The more detailed process of orientated growth is presented as Algorithm ~\ref{alg:alg2}.

\begin{algorithm}
\caption{Process of Orientated Growth}
\label{alg:alg2}
    \begin{algorithmic}
        \STATE
        \STATE {\textbf{Input:} \text{rough connected component $T_r$.}}
        \STATE {\textbf{Output:} \text{complete connected component }$T_c$.}
        \STATE {\text{Determine centroid $c_0$ and initial slope $k_0$, $L=0$.}}
        \STATE {\text{Search for the optimal $k$ in the neighborhood of $k_0$.}}
        \STATE {\text{Determine the central axis layer $m$ and the seed $S$.}}
        \WHILE{not all directions traversed}
        \STATE Calculate $p_\mathrm{all\_t}$ and $p_\mathrm{all\_b}$ of newly attached pixels $P_n$.
        \IF{$p_\mathrm{all\_t} > p_\mathrm{all\_b}$ and $L \leq L_\mathrm{max}$}
        \STATE $S:=S\cup P_n$
        \STATE $L := L+1$
        \ELSE
        \STATE $L:=0$
        \ENDIF
        \ENDWHILE
        \STATE Return $S$ as $T_c$.
    \end{algorithmic}
\end{algorithm}
\section{Data Source and Parameter Selection}
\subsection{Train Set Establishment}
\begin{figure}
    \centering
    \subfloat[]{\includegraphics[width=\columnwidth]{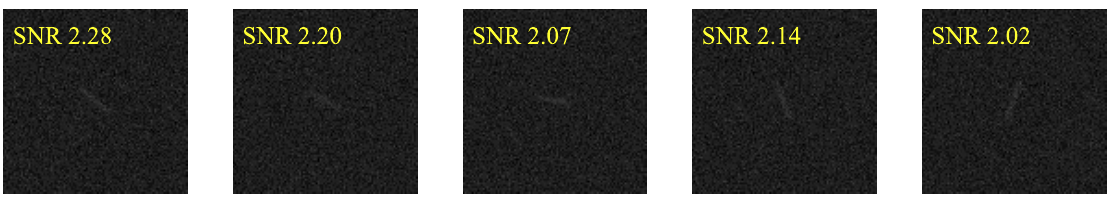}}\\
    \subfloat[]{\includegraphics[width=\columnwidth]{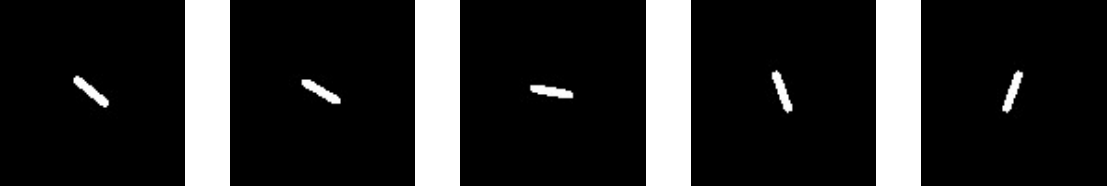}}
    \caption{Some samples from train set. (a) Raw images of the samples. (b) Ideal connected components of the samples}
    \label{fig:16}
\end{figure}

Firstly, streaks with random directions and lengths are generated on a Gaussian noise background, where the noise $g_N \sim N(30,64)$, the angle of streaks $\theta\sim U(0,180)$ and the length of streaks $l\sim U(10,22)$. Let the center of the streak be $(x_c, y_c)$, then the central axis of streak can be described as 
\begin{equation}
    l_s:y=k(x-x_c)-y_c
\end{equation}
where $k=\tan\theta$ denotes the slope of the streak and $x$ is within the range of $[x_c-\frac{1}{2}l\cos\theta, x_c+\frac{1}{2}l\cos\theta]$. The intensity of the streak follows the formula (1), and the PSNR is set to be around 2 via the selection of $I_c$ and $\sigma$. Some raw images in the train set and their ideal target regions are shown in Fig.~\ref{fig:16}.

The ideal target region is obtained by thresholding, and the threshold is selected to 4. A low threshold is adopted to improve the recall ratio of the trained SVC, as a higher threshold may conduce to the fracture of connected components. As the proportion of the streak area in the whole image is very small, an adaption needs to be performed for sample balance. In addition, the proportion of edge pixels is also increased for further improvement of classification capability. Finally, background pixels and target pixels take 60 percent and 40 percent respectively, and the train set contains more than 130,000 samples with 26 dimensions. 

\subsection{Threshold Determination}
The output of the SVC is determined directly by the threshold. A lower threshold will expand connected components and more redundant pixels will be included, while a higher threshold will shrink connected components and missing pixels will increase. Both cases can deteriorate the quality of connected components, and based on experiments, the threshold $T_h$ in the proposed algorithm is chosen as 0.5. 

\subsection{Hardware Specification}
\begin{figure}
    \centering
    \includegraphics[width=0.9\columnwidth]{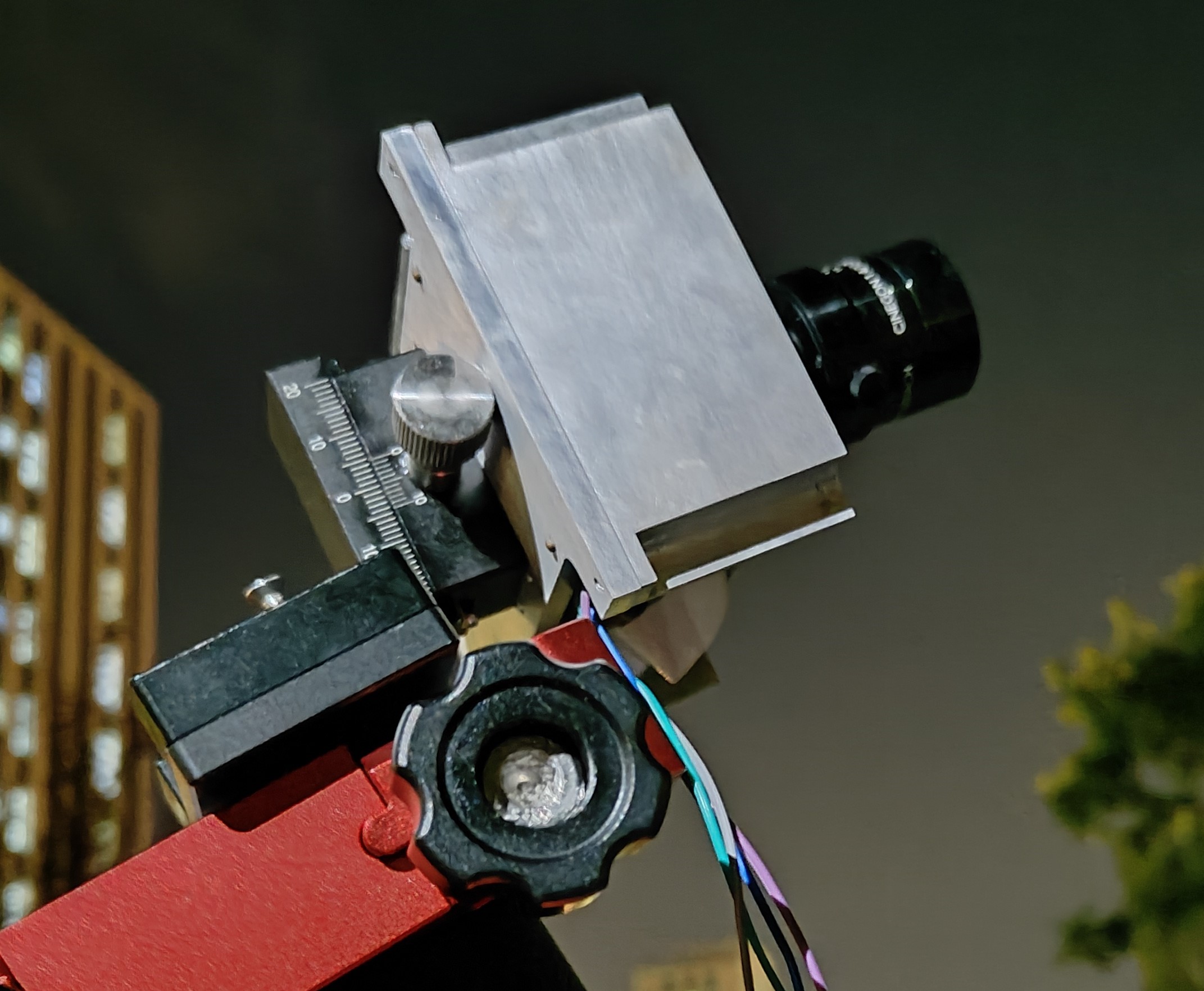}
    \caption{The star tracker used in this paper}
    \label{fig:hardware}
\end{figure}

The hardware platform involved in this paper is mainly for simulation and image acquisition. Simulation is accomplished via MATLAB2019b and PC platform with Intel i7-11800H CPU, while image acquisition is achieved with a star tracker that photographs the celestial sphere. The star tracker adopted in experiments has a focal length of $25mm$, a FoV of $25^\circ$, a resolution of $2048\times2048$, and a unit scale of $5.5\mu m$. The photo of image acquirement with the star tracker adopted is shown in Fig.~\ref{fig:hardware}.

\section{Experiments and Result}
The minimum extractable SNR and centroid error are adopted to evaluate the capacity and precision of the proposed algorithm, respectively. The centroid $(x_c, y_c)$ is defined as below:
\begin{gather}
    x_c=\frac{\sum_i x_i g_i}{\sum_i g_i}\\
    y_c=\frac{\sum_i y_i g_i}{\sum_i g_i}
\end{gather}
With the assumption that the reference centroid is $(x_0, y_0)$, the absolute centroid error can be expressed as 
\begin{equation}
    err=\sqrt{(x_c-x_0)^2 + (y_c-y_0)^2}
\end{equation}
\subsection{SVC Performance Analysis}
\begin{figure}
    \centering
    \includegraphics[width=0.7\columnwidth]{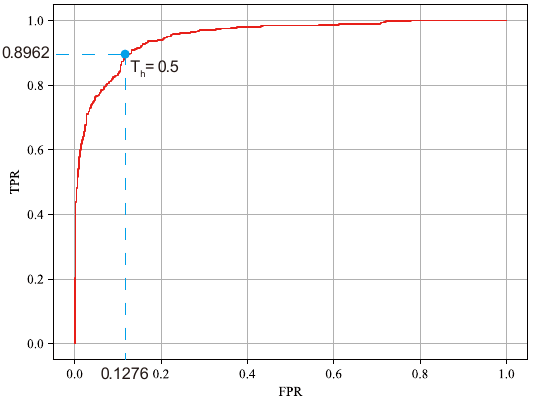}
    \caption{The ROC of the trained SVC}
    \label{fig:17}
\end{figure}

For model validation, the K-Fold method is performed and the $K$ is selected to 5. The accuracy of each group and the average accuracy are shown in the table. ~\ref{tab:1}:
\begin{table}[]
    \centering
    \caption{The accuracy of the SVC in each test.}
    \begin{tabular}{|c|c|c|c|c|c|c|}
    \hline
    \textbf{Group} & \textbf{1} & \textbf{2} & \textbf{3} & \textbf{4} & \textbf{5} & \textbf{Average}\\
    \hline
        Accuracy/\% &  89.82 & 90.23 & 91.02 & 88.98 & 89.13 & 89.84\\
    \hline
    \end{tabular}
    \label{tab:1}
\end{table}
The ROC of the trained SVC is shown in Fig.~\ref{fig:17}.

According to the curve, the possibility of classifying a background pixel as the target is 0.1267 with a threshold of 0.5. But it doesn’t mean every background pixel shares the same possibility, as most of the misclassified pixels are generated by the dilation effect of the template and can be ascribed to the influence of the target. Other pixels of misclassification are generated by Gaussian noise in the background, but they can hardly form connected components that are large enough. Since a filter of connected component size is utilized, the latter type of misclassified pixels will not affect results significantly. 

\begin{figure}
    \centering
    \includegraphics[width=0.8\columnwidth]{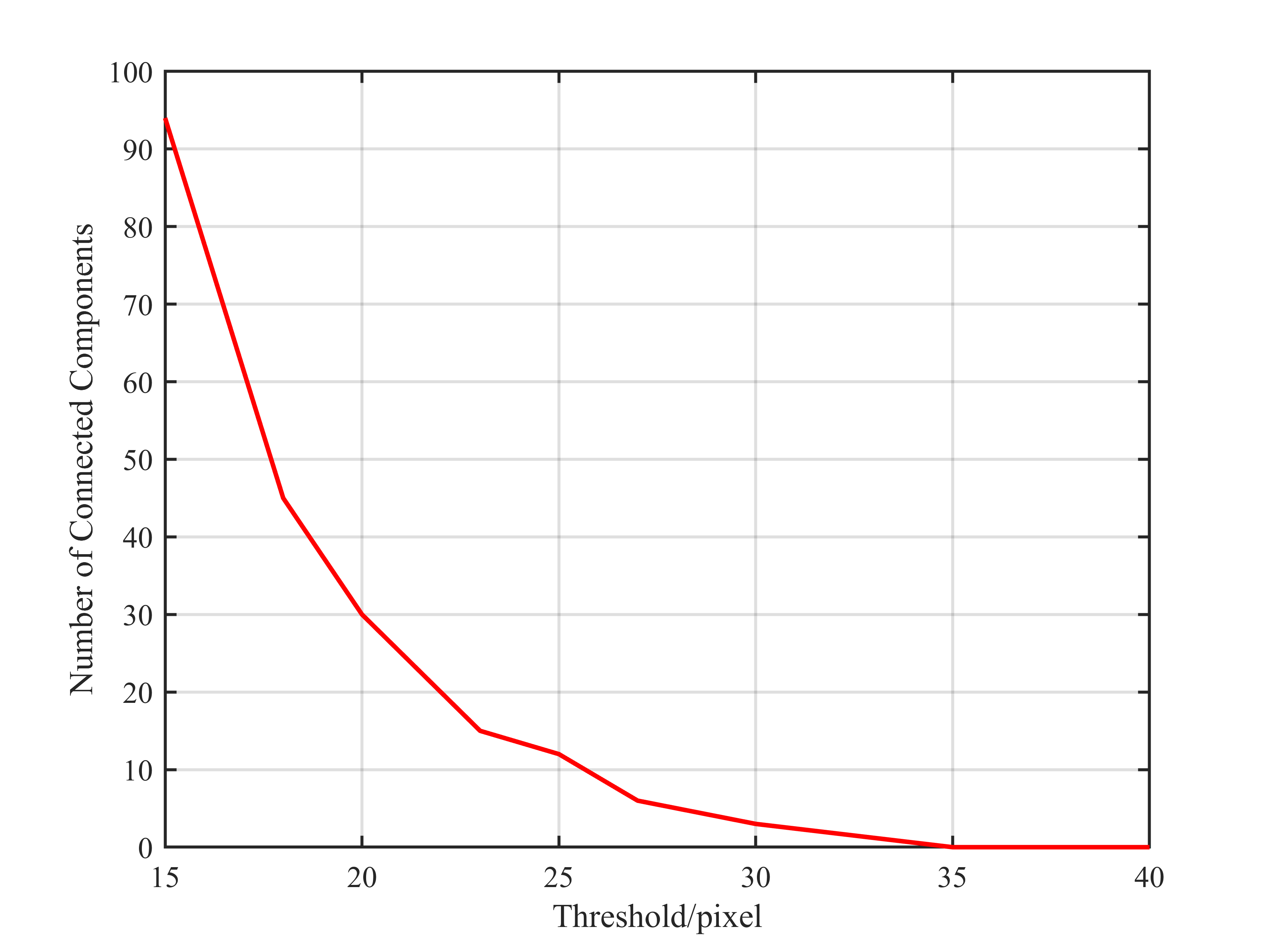}
    \caption{The curve of the number of false targets changing with the threshold}
    \label{fig:18}
\end{figure}

Then a pure background with a resolution of $1024\times1024$ is generated and processed by the proposal algorithm for verification. Fig.~\ref{fig:18} shows the relationship between the number of fake targets and the threshold of the connected component size. 

\begin{figure}
    \centering
    \subfloat[]{\includegraphics[width=\columnwidth]{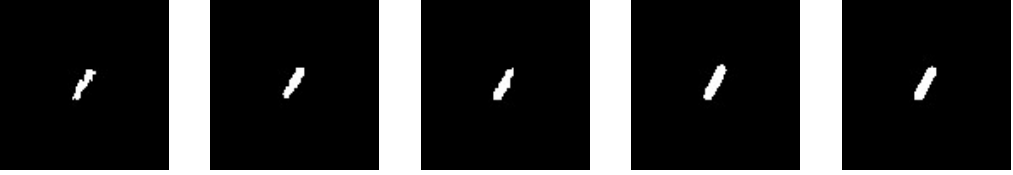}}\\
    \subfloat[]{\includegraphics[width=\columnwidth]{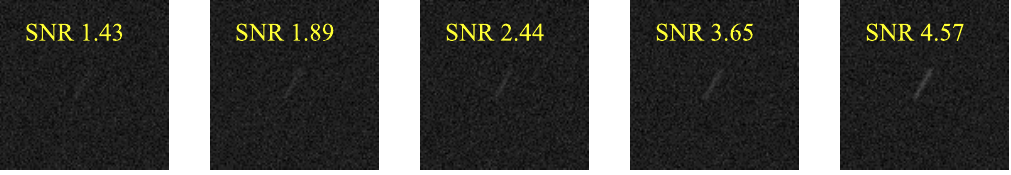}}
    \caption{(a) Results of crude classification. (b) Raw images of streak targets with different PSNR}
    \label{fig:19}
\end{figure}

As shown in Fig.~\ref{fig:18}, the number of fake targets declines with the threshold increasing, and no fake target is detected when the threshold is larger than 35. After verifying the effectiveness of the SVM model, the research also studied the effect of the crude classification method on target extraction with different signal-to-noise ratios. Streaks are generated with a fixed length of 20 pixels and a direction of $60^\circ$ and the results are shown in Fig.~\ref{fig:19}.

\begin{figure}
    \centering
    \includegraphics[width=0.8\columnwidth]{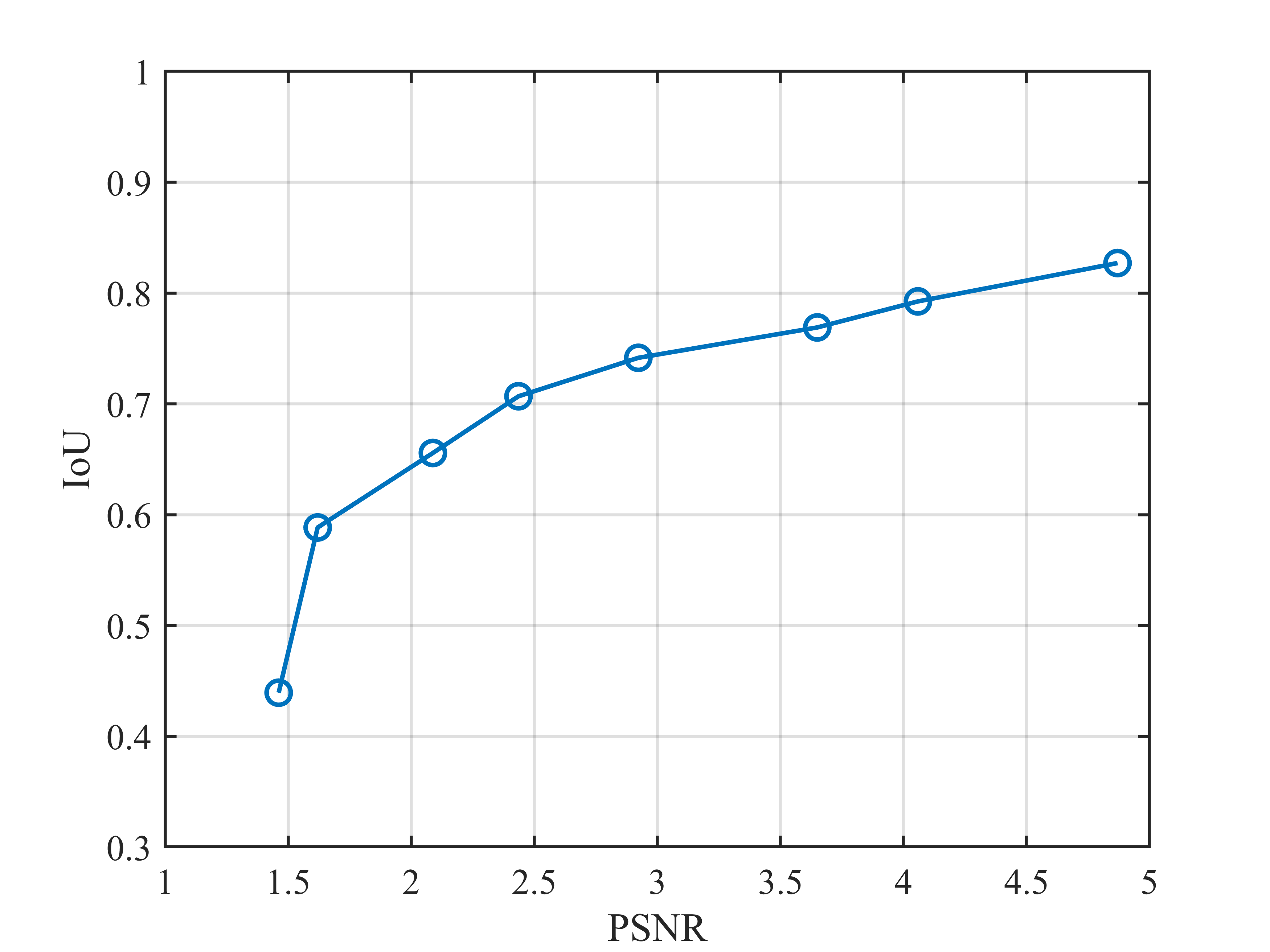}
    \caption{Curve of the relationship between the IoU and the PSNR}
    \label{fig:20}
\end{figure}

Consistent with intuition, the edges of the connected components can be smoother under the conditions of higher SNR, and the shapes of the results are closer to the reference streaks. In addition, the IoUs of the crude classification results and the ideal streaks under different PSNRs were also calculated, as shown in Fig.~\ref{fig:20}.

From Fig.~\ref{fig:20}, as the PSNR increases, the IoU constantly increases with the slope declining, and it is above 0.75 while the PSNR is relatively high. In the case of extremely low PSNR, this trend will help the crude classification result to contain more effective pixels, so the IoU will grow faster. In the general case of low PSNR, this trend also includes some non-target pixels while expanding the connected components, which leads to a slowdown in the growth rate of IoU. Although the SVC in this paper is trained based on targets with a PSNR around 2.0, the classifier can still effectively recognize streaks with an SNR as low as 1.4.

\subsection{Results of Orientated Growth}
Orientated growth will be performed based on the crude classification results, and the method will be compared with the crude classification and other commonly used connected component extraction methods, such as the ODCC algorithm. For this purpose, multiple sets of experiments are designed to test the performance of the algorithm under different conditions.
\subsubsection{Method Verification}
\begin{figure}
    \centering
    \subfloat[PSNR=1.44]{\includegraphics[width=0.8\columnwidth]{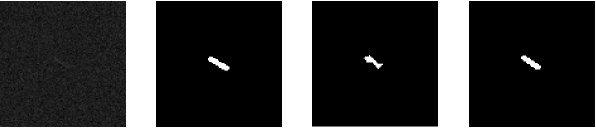}}\\
    \subfloat[PSNR=3.68]{\includegraphics[width=0.8\columnwidth]{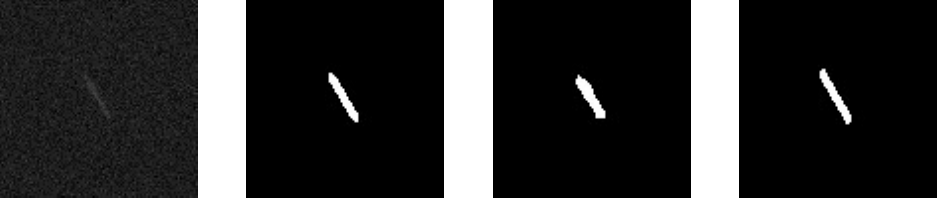}}
    \caption{Comparison of results of crude classification and orientated growth with different PSNRs and four images in each group denote raw images of targets, ideal connected components, results of crude classification, and results of orientated growth, respectively.}
    \label{fig:21}
\end{figure}

To verify the proposed method, two streaks with PSNRs of 1.44 and 3.68 are generated and processed by crude classification and orientated growth subsequently. The results are shown in Fig.~\ref{fig:21}.

\begin{figure}
    \centering
    \subfloat[PSNR=1.44]{\includegraphics[width=0.415\columnwidth]{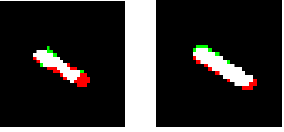}}
    \quad
    \subfloat[PSNR=3.68]{\includegraphics[width=0.4\columnwidth]{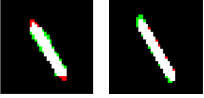}}
    \caption{Comparisons of connected components before and after orientated growth.}
    \label{fig:22}
\end{figure}

Then the connected components of crude classification and orientated growth are compared with the ideal connected component respectively in Fig.~\ref{fig:22}.

It can be seen from the above comparisons that when the PSNR of the target is extremely low, the shape of the connected component obtained by crude classification is quite different from the ideal connected component, not only there are missing endpoints, but also the width is significantly narrower. Even if the PSNR increases, the connected component extracted by crude classification still has the problem of missing endpoints and is slightly wider than the ideal connected component. After orientated growth, the result almost coincides with the reference streak, and the complete reconstruction of targets can be achieved.

To further understand the influence of orientated growth on centroid errors, the proposed algorithm and the ODCC method are performed under different PSNRs to extract space objects of the same length. Each result is calculated by an average centroid error of 200 stars, and the experiments in the following chapters are all carried out in this form.

\begin{figure}
    \centering
    \includegraphics[width=0.8\columnwidth]{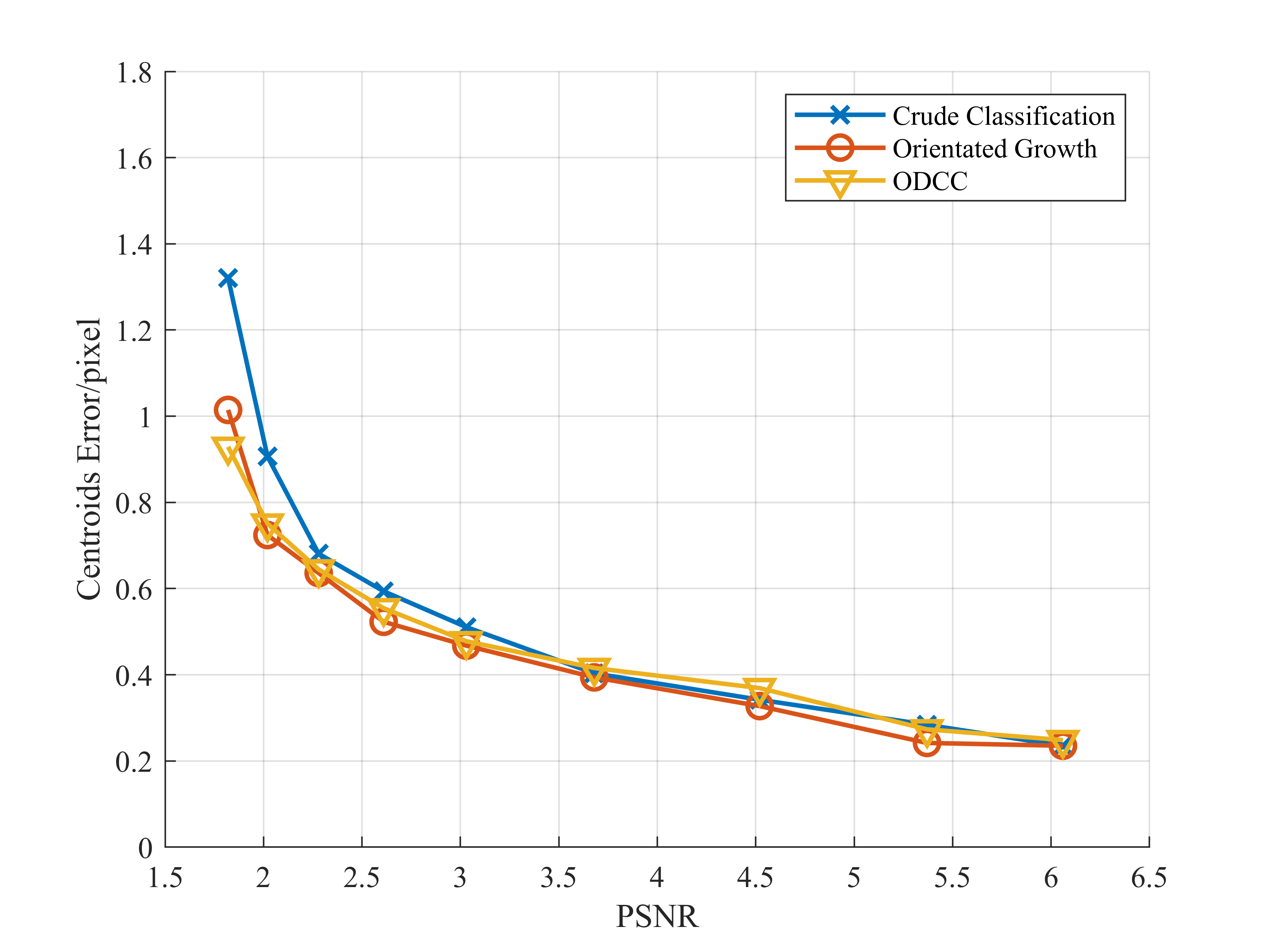}
    \caption{Centroid errors of streaks extracted at different PSNRs.}
    \label{fig:23}
\end{figure}

As the curve shown in Fig.~\ref{fig:23}, with the increase of the PSNR, the accuracy of the proposed method is continuously improved, and the centroid errors of the orientated growth and crude classification are getting close since the increasing PSNR leads to a higher IoU. Besides, it can be observed that under the condition of extremely low SNR, the precision of the method in this paper is slightly lower than that of the ODCC method, which is caused by the quality degradation of the connected components. While the PSNR is higher, the curve proves that the optimal solution obtained by the local search in the proposed method is almost equivalent to that obtained by the global search in the ODCC method.

\subsubsection{Extraction Ability Test}
\begin{figure}
    \centering
    \subfloat[]{\includegraphics[width=\columnwidth]{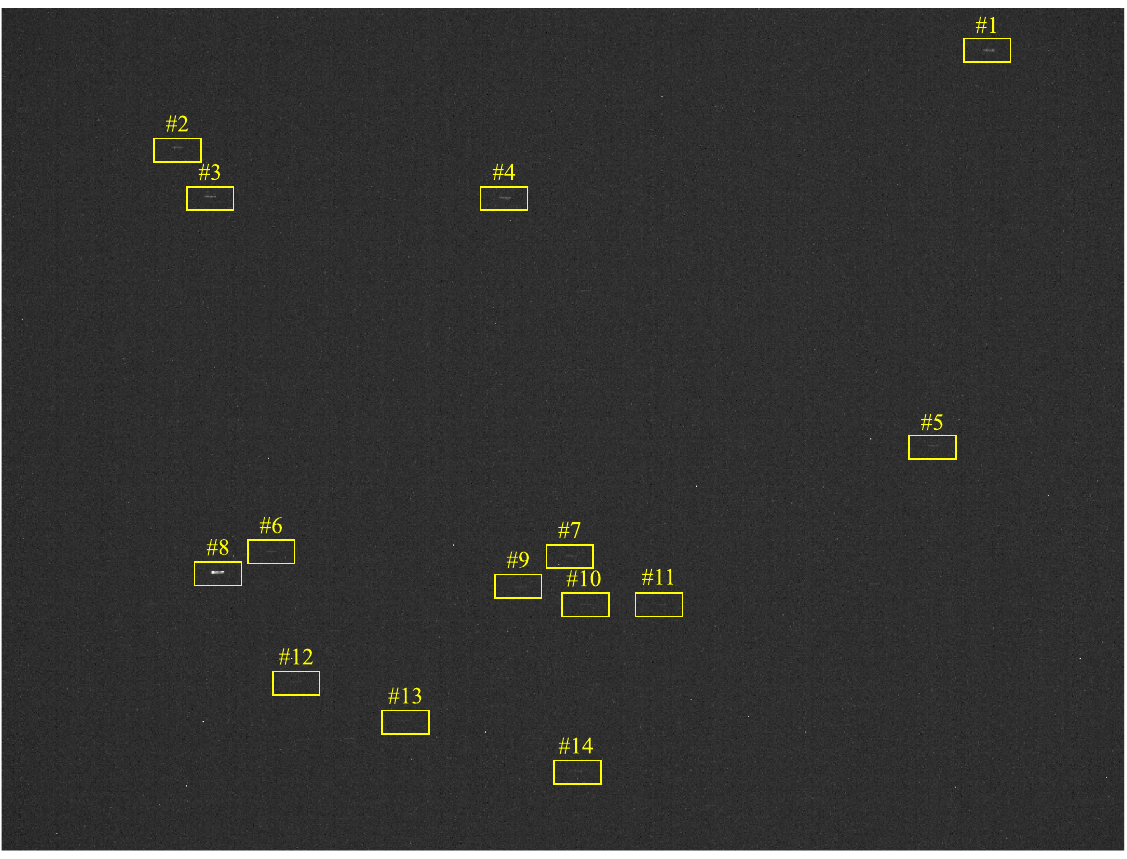}}\\
    \subfloat[]{\includegraphics[width=\columnwidth]{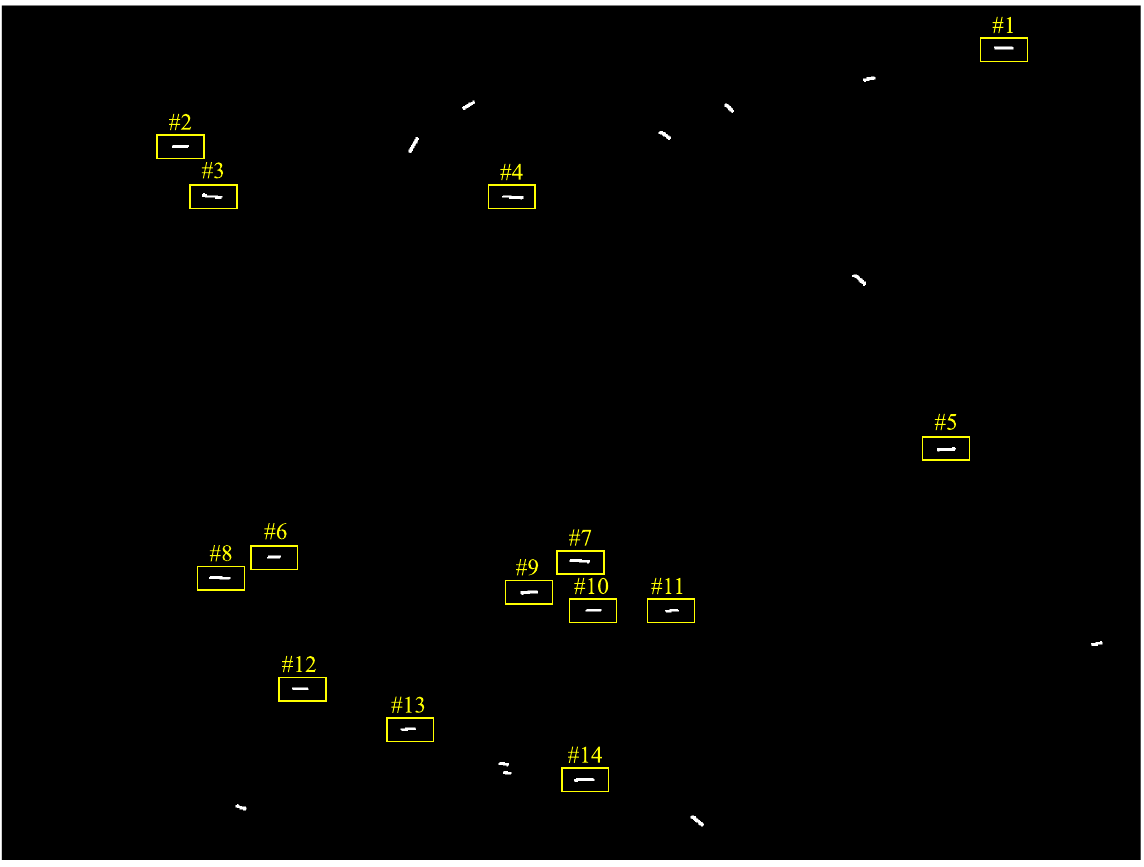}}\\
    \subfloat[]{\includegraphics[width=\columnwidth]{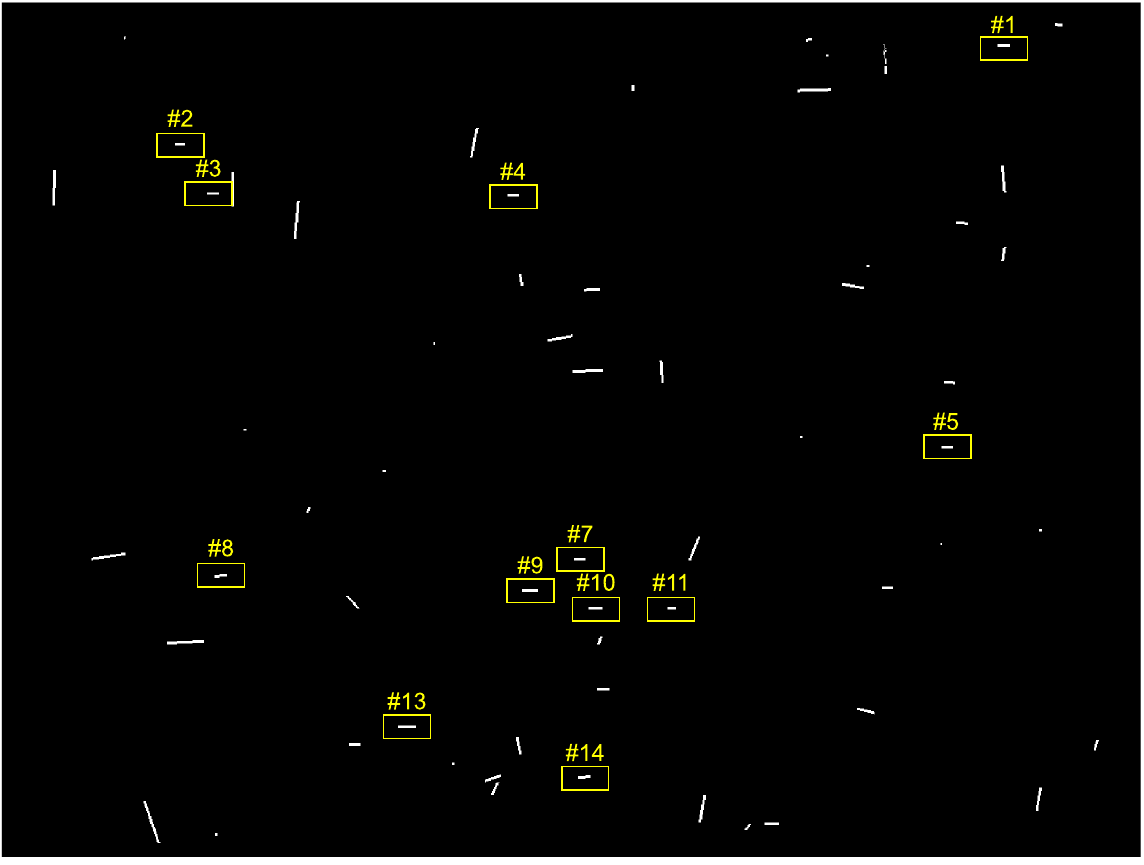}}
    \caption{(a) Raw image obtained by star tracker. (b) Connected components returned by the orientated growth algorithm. (c) Connected components returned by the ODCC algorithm}
    \label{fig:24}
\end{figure}

\begin{figure}
    \centering
    \includegraphics[width=0.8\columnwidth]{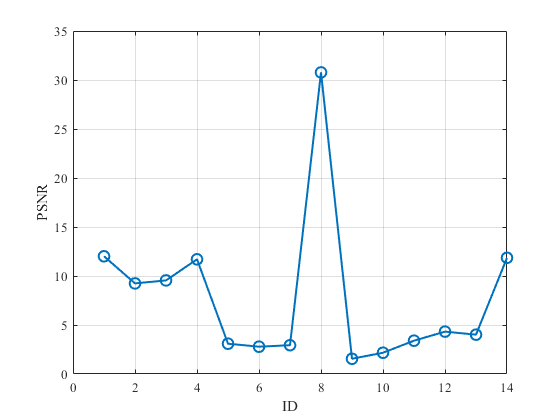}
    \caption{PSNRs of star streaks in the obtained image.}
    \label{fig:25}
\end{figure}

A star tracker is used to shoot star maps under the conditions of $10^\circ/s$ angular velocity and $20ms$ exposure time. Then the method of this paper is applied to extract the streaks from the star tracker image. Finally, a comparison is performed with the extraction results of the ODCC method, as shown in Fig.~\ref{fig:24}, and the PSNRs of streaks in the processed image are shown in Fig.~\ref{fig:25}.

It can be found that there are many noise points and strip fixed pattern noise (FPN) in the raw image, and the PSNR of the star points reaches 30.77 at the highest and 1.51 at the lowest. According to the above results, the proposed algorithm can overcome the adverse influence of the background noise, and effectively extract star streaks with various PSNRs. In addition, compared with the ODCC, the proposed algorithm can extract more star points, and there are fewer wrong results, which represents higher robustness for different noise patterns. When looking into the extraction results of the ODCC algorithm, the directions of the fake streaks are mostly horizontal or vertical, as the ODCC method enhances streaks and stripe noises existing in the raw image simultaneously. In contrast, the proposed method utilizes the difference between the gray distribution pattern of the target and its neighborhood, and the impact of the strip noise will be effectively suppressed by averaging, so the algorithm in this paper has a stronger tolerance to strip noise.

\subsubsection{Precision under streaks with different lengths}
\begin{figure}
    \centering
    \subfloat[]{\includegraphics[width=0.5\columnwidth]{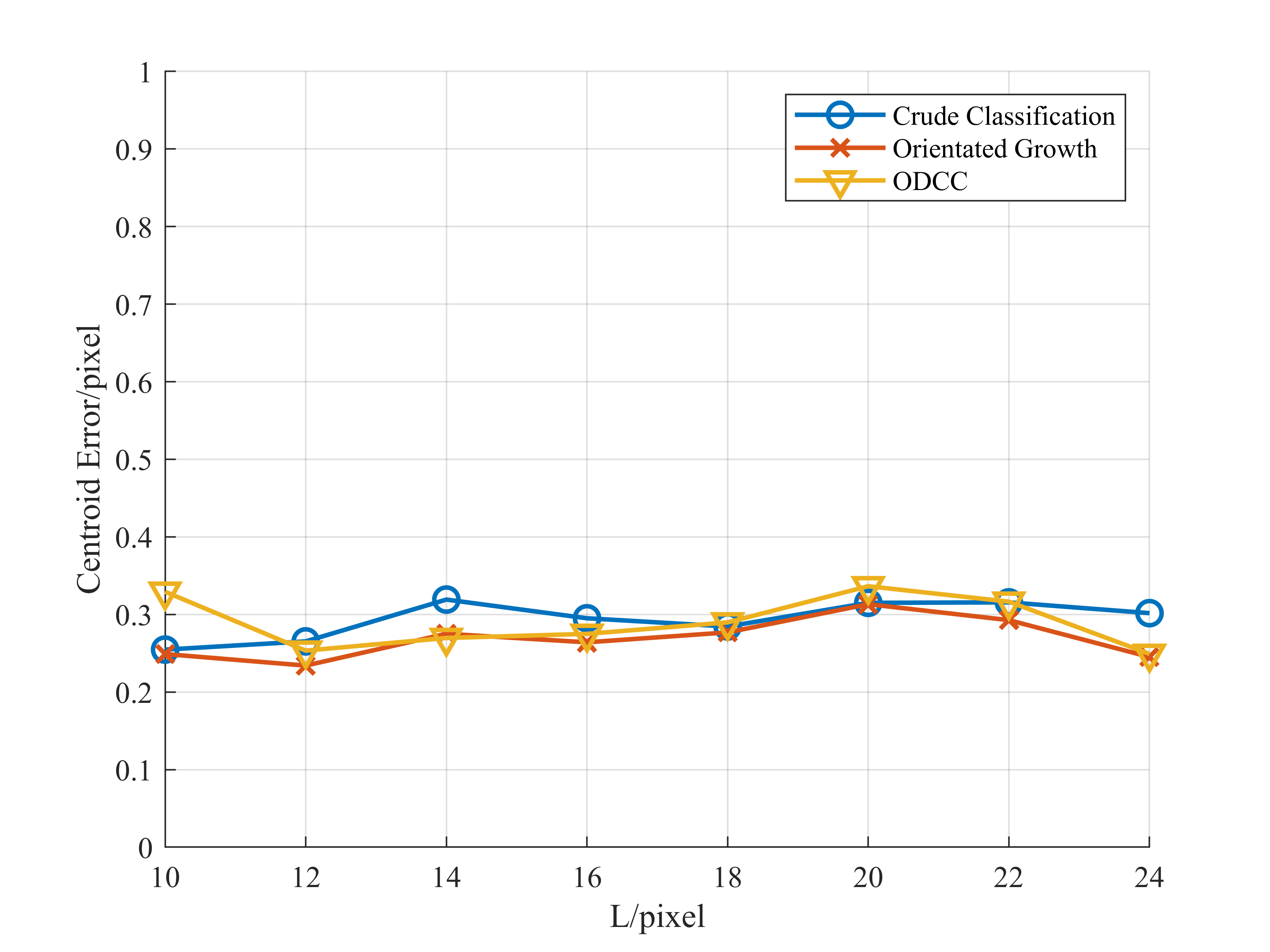}}
    \subfloat[]{\includegraphics[width=0.5\columnwidth]{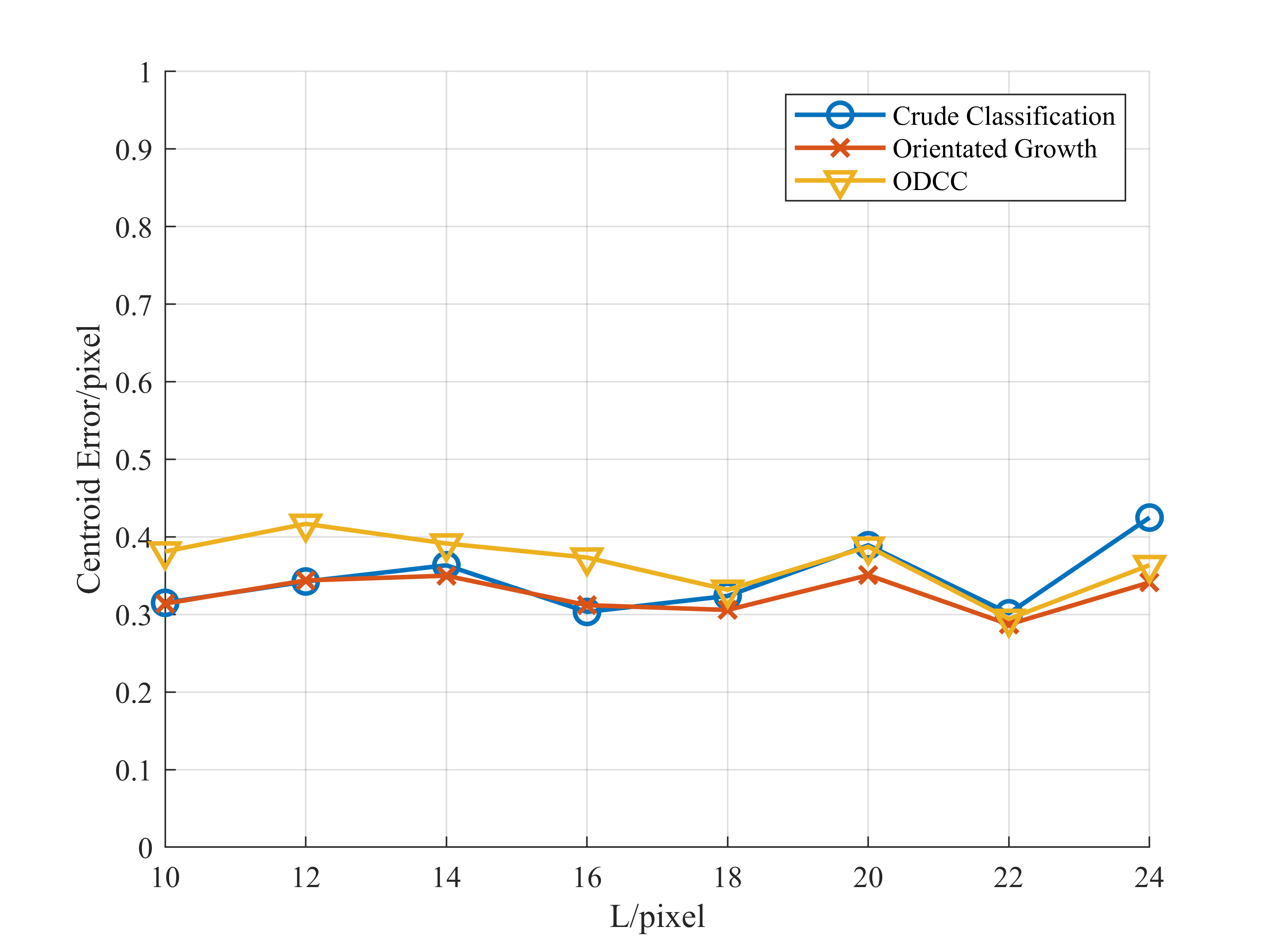}}\\
    \subfloat[]{\includegraphics[width=0.5\columnwidth]{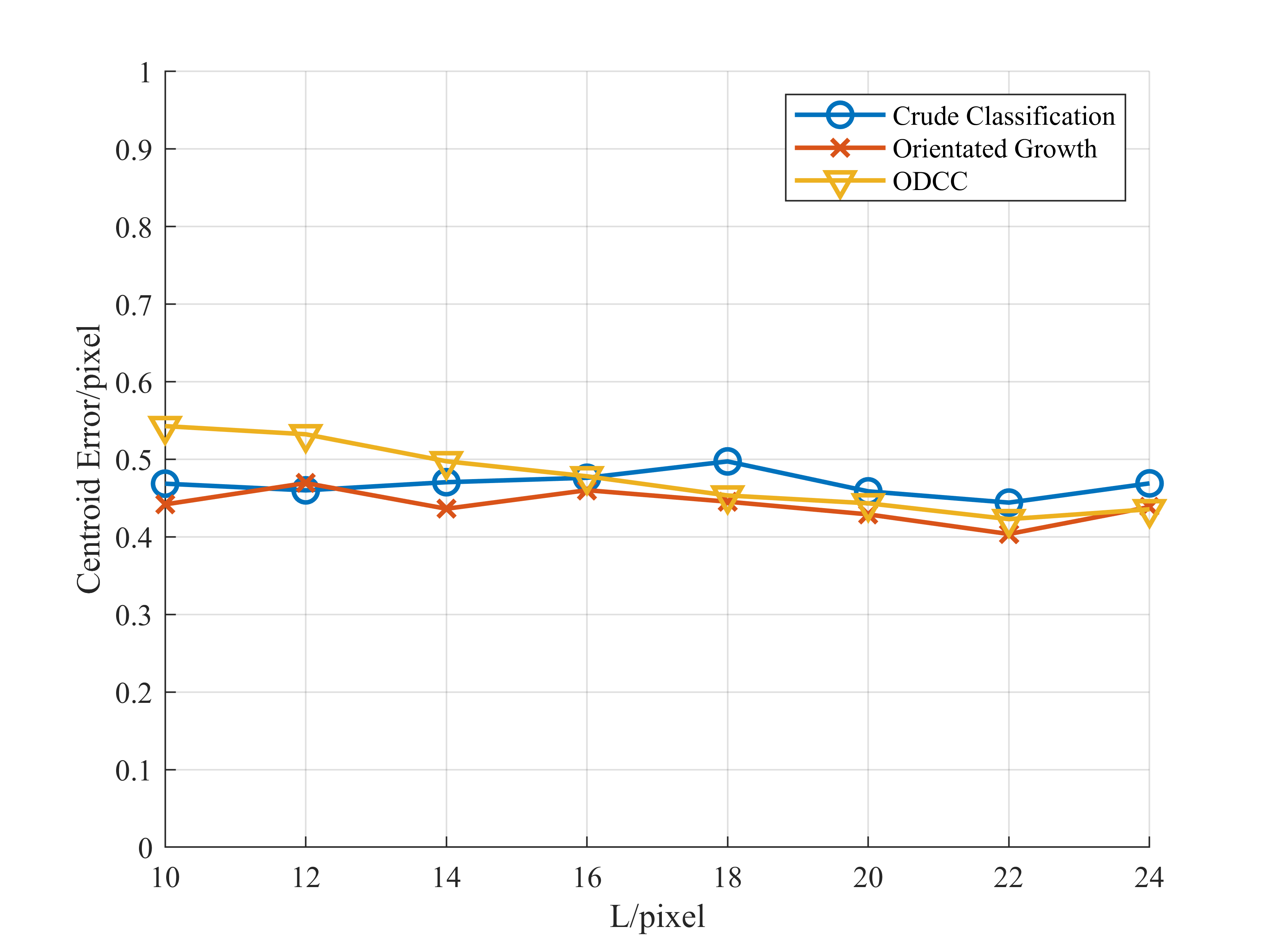}}
    \subfloat[]{\includegraphics[width=0.5\columnwidth]{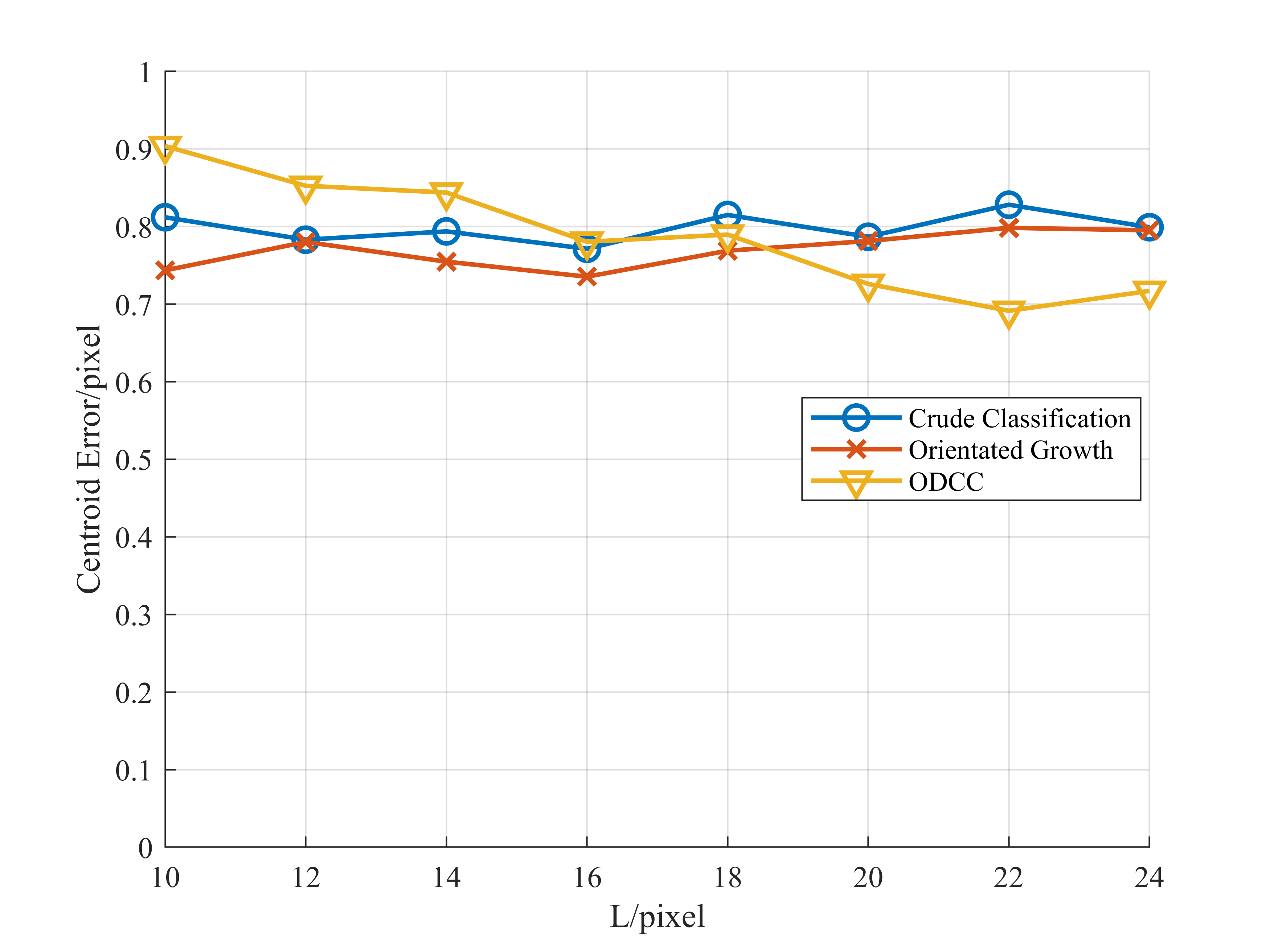}}
    \caption{Centroid errors of extractions under different streak lengths at (a) PSNR 5.0, (b) PSNR 4.0, (c) PSNR 3.0, and (d) PSNR 2.0, respectively.}
    \label{fig:26}
\end{figure}

Under the conditions of different PSNRs, the proposed algorithm and the ODCC algorithm are used to extract streaks of different lengths, and the centroid error is calculated respectively, as shown in Fig.~~\ref{fig:26}. According to the curves, the centroid errors of the two methods are close when the PSNR is relatively high, and the centroid errors of both algorithms manifest the resemble tendency while PSNR changes. Besides, the algorithm proposed in this paper presents less sensitivity to the length of streaks, while the precision of the ODCC method in short streak extraction has a significant degradation. The size of the convolution template adopted in the detection stage of the ODCC is responsible for such degradation, as it determines the lower limit of the extractable length directly. In addition, when it comes to longer streaks at extremely low PSNR, the performance of the proposed method is not as good as that of the ODCC method. The reason is that defects like fractures will occur more frequently in longer streaks, and there is also an upper limit to the length of orientated growth, so the missing part may not be complemented.

\subsubsection{Precision under different deviations of background noise}
\begin{figure}
    \centering
    \includegraphics[width=0.7\columnwidth]{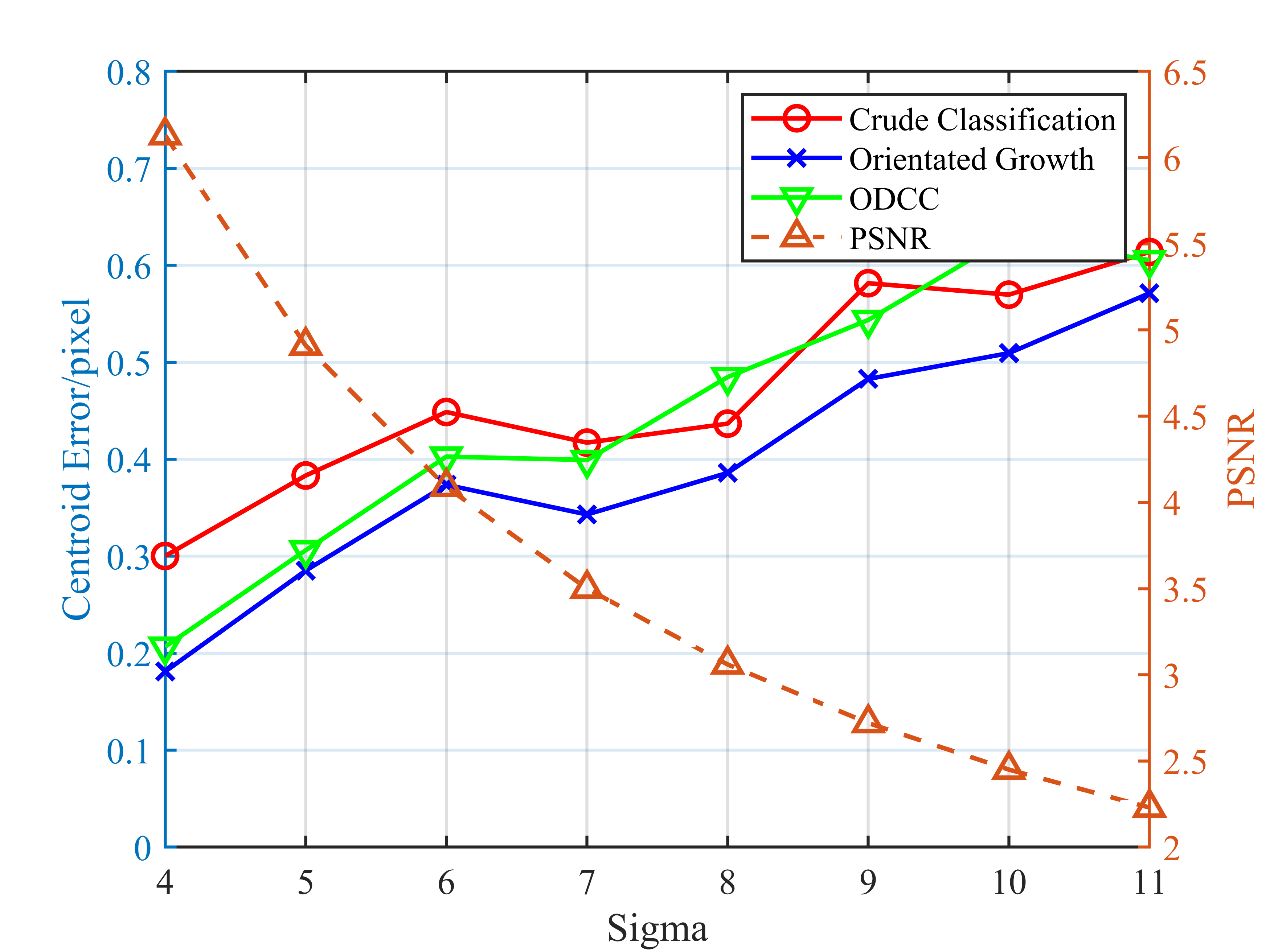}
    \caption{Centroid errors under different standard deviations of background noise.}
    \label{fig:27}
\end{figure}

In this experiment, the PSNR of the target is altered by changing the standard deviation of the background noise, and the length of the streaks is set to 16 pixels. The proposed algorithm and the ODCC method are performed respectively, and the results are shown in Fig.~\ref{fig:27}.

It can be seen from the figure that as the standard deviation of the background noise increases, the PSNR of the target decreases, and the precision of the extracted centroid decreases as well. This tendency is consistent with the previous analysis. It is worth noting that although the feature weights of the algorithm in this paper are obtained based on background noise $g_n\sim N(30,64)$, the SVC can adapt to the background of different standard deviations without retraining, which indicates good robustness. In conclusion, when compared with ODCC, the proposed algorithm can achieve higher precision of target extraction with the background of different standard deviations.

\subsubsection{Efficiency compared with the ODCC}
In terms of processing speed, the method in this paper has great advantages over the ODCC. During the experiment, the original image with a resolution of $1280\times960$ was processed 150 times by the method in this paper and the ODCC method respectively based on MATLAB2019, and the average processing time was then calculated. The proposed algorithm takes an average time of 0.1645s per frame, while ODCC takes an average of 1.1936s. The algorithm in this paper is significantly more efficient. While looking into the algorithm flows of the two, it can be found that the ODCC method adopts 15 different directional templates for filtering in the detection stage and then performs a nonmaximal suppression process for each filtered result, which greatly increases the calculation time. In contrast, the proposed algorithm only performs one filtering in the detection stage and does not involve non-maximum suppression to achieve preliminary target segmentation. As a result, the calculation process is more simplified, and the extraction effect is consistent with ODCC.

\section{conclusion}
The increasing space debris population has posed an increasing threat to the safety of spacecraft. To improve the space situational awareness (SSA) of spacecraft, it is of great significance to research space debris detection methods based on space-based cameras. The high speed and the low reflect intensity of space debris hurdle the extraction with traditional methods; and due to the limitation of satellite hardware resources, complex algorithms can hardly be implemented. To solve the above problems, this paper proposes an extraction method for high-dynamic space objects based on local contrast and orientated growth of connected components.
Firstly, a linear template based on the idea of local contrast is proposed for detection in this paper, and the SVM with a linear kernel determines the feature weights in the template. Then, to include as many target pixels as possible, apart from the shape and direction constraints of the streak, MLE is also utilized to realize the orientated growth of the target connected component, which guarantees the complete extraction of the target and avoids inefficient global searches. From the test results, when the PSNR of the spatial objects extracted by the proposed method is about 2, the average centroid error does not exceed 1 pixel, which is consistent with the theoretical optimal method like ODCC.

\section{Further Discussion}
The ODCC method filters the images with a convolution template to extract the direction of the target and then performs a global search for the optimal connected component based on the MLE. Therefore, the target connected components returned by this method are optimal to some extent. In contrast, this paper does not involve a convolution template, and while reconstructing connected components, only local searches are performed based on the crude classification results. According to the outcomes, the proposed method can still achieve the same or even higher precision as the ODCC method, and a significantly higher efficiency can also be reached. For an explicit explanation, further research on the proposed algorithm is proceeded.

\begin{figure}
    \centering
    \includegraphics[width=0.8\columnwidth]{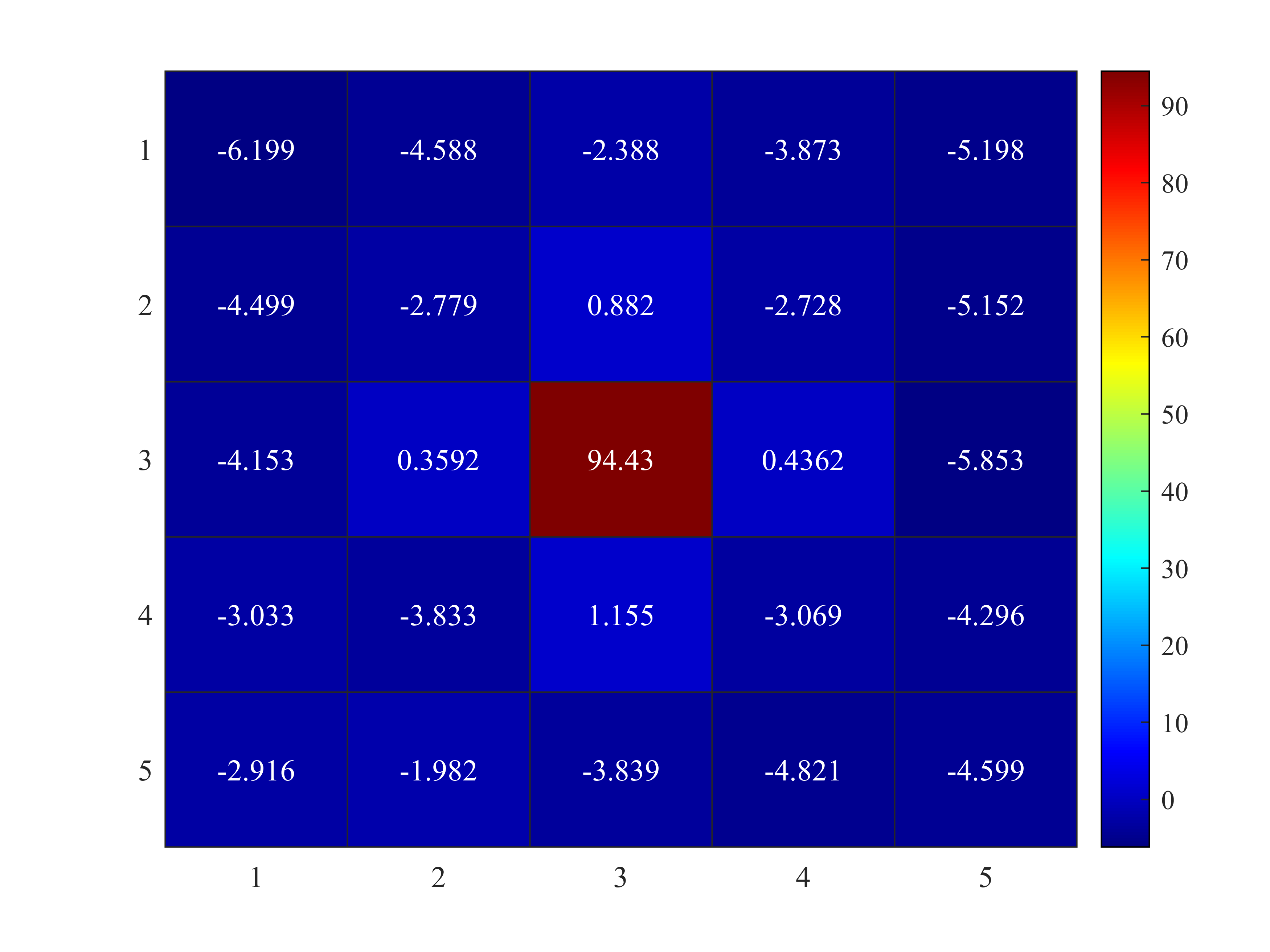}
    \caption{Heatmap of the SVC weight matrix}
    \label{fig:29}
\end{figure}

The research first concentrates on the template adopted in the crude classification stage. As mentioned above, the weight vector consists of 26 elements, whose first 25 elements denote the average gray of each sub-region. These feature weights are shown below in the form of a heatmap as Fig.~\ref{fig:29}.

The heatmap intuitively depicts that the average gray of the target sub-region corresponds to a much higher weight than other areas, and the farther away from the center, the lower the weight, showing the characteristics of a near Gaussian distribution. This is because although the proposed template is designed for streaks, as the target trajectory can be located at the template arbitrarily, the weight matrix returned in the end cannot show a strong directionality but should be isotropic. Thus, while classifying with SVC, the weighted sum of the first 25 features and the SVC weights is equivalent to the result of performing a Gaussian template convolution filtering. When there is a streak passes through the target sub-region, only a few sub-regions are affected by the target due to the large template, and most of the sub-regions can be considered as simple backgrounds. The distributions of mean features in the template roughly compose a pattern with high middle and low sides, that is, a Gaussian-like distribution. Therefore, while filtering with a near Gaussian template, the existence of targets will contribute to a higher response. In addition, although the template proposed in this paper is mainly used for dim streak detection, the weight matrix endows the template with the ability to detect targets with various shapes, such as dim points. Compared with the ODCC method, the proposed detection method has a wider application scenario.

Apart from the average gray of the target sub-region, the target area has other salient features, so the maximum gray of the target region is introduced as the $26$th feature, and the classification process can be expressed as follows:
\begin{equation}
    y=\sum_{i=1}^25\omega_ix_i+\omega_{26}x_{26} + b
\end{equation}

Arbitrarily, if the classification threshold is set to $t_0$, and in the case of $y\geq t_0$,  then the above formula can be written as 
\begin{equation}
    x_{26}\geq\frac{t_0-b}{\omega_{26}}-\frac{1}{\omega_{26}}\sum_{i=1}^{25}\omega_ix_i
\end{equation}

The formula (39) reveals that the value of $x_{26}$ determines the result of the classification to some extent, as only when the highest gray of the target sub-region is greater than a specific value, the central pixel of the template will be classified as the target pixel. Besides, as Fig.~\ref{fig:8} shows, the $\omega_{13}$ and $\omega_{26}$ are the highest and the second highest weights of the weight vector, which indicates that the features of the target sub-region pose decisive influence on classification results.

When it comes to the template design, if we reduce the scale of the sub-regions in the template except the target domain to obtain a feature space with more dimensions, or in other words, increase the sampling rate of the mean feature, then theoretically, the weight matrix will be closer to the two-dimensional Gaussian distribution, and the detection effect is at least as good as the current method. However, on the premise of using the linear kernel SVC and the classification accuracy of the proposed method is close to 90\%, the further improvement of the accuracy will be significantly affected by the edge effect. Actually, not only the algorithm performance can be barely improved, but also the computational complexity will increase greatly. Similarly, it comes to the same situation when expanding the template, because of the edge effect.

 While detecting with a smaller template, the number of sub-regions that are affected by the streak will increase consequently, resulting in the enlargement of the difference between the Gaussian distribution and the average gray distribution within the template. Additionally, as fewer features can be extracted from a smaller template, the accuracy of the classifier will be reduced significantly. Above all, a trade-off should be made to achieve the balance between computation and effect, and the proposed template is the optimum under the conditions of this paper.

\bibliographystyle{IEEEtran}

\bibliography{main}

\vspace{-10mm}
\begin{IEEEbiography}[{\includegraphics[width=1in,height=1.25in,clip,keepaspectratio]{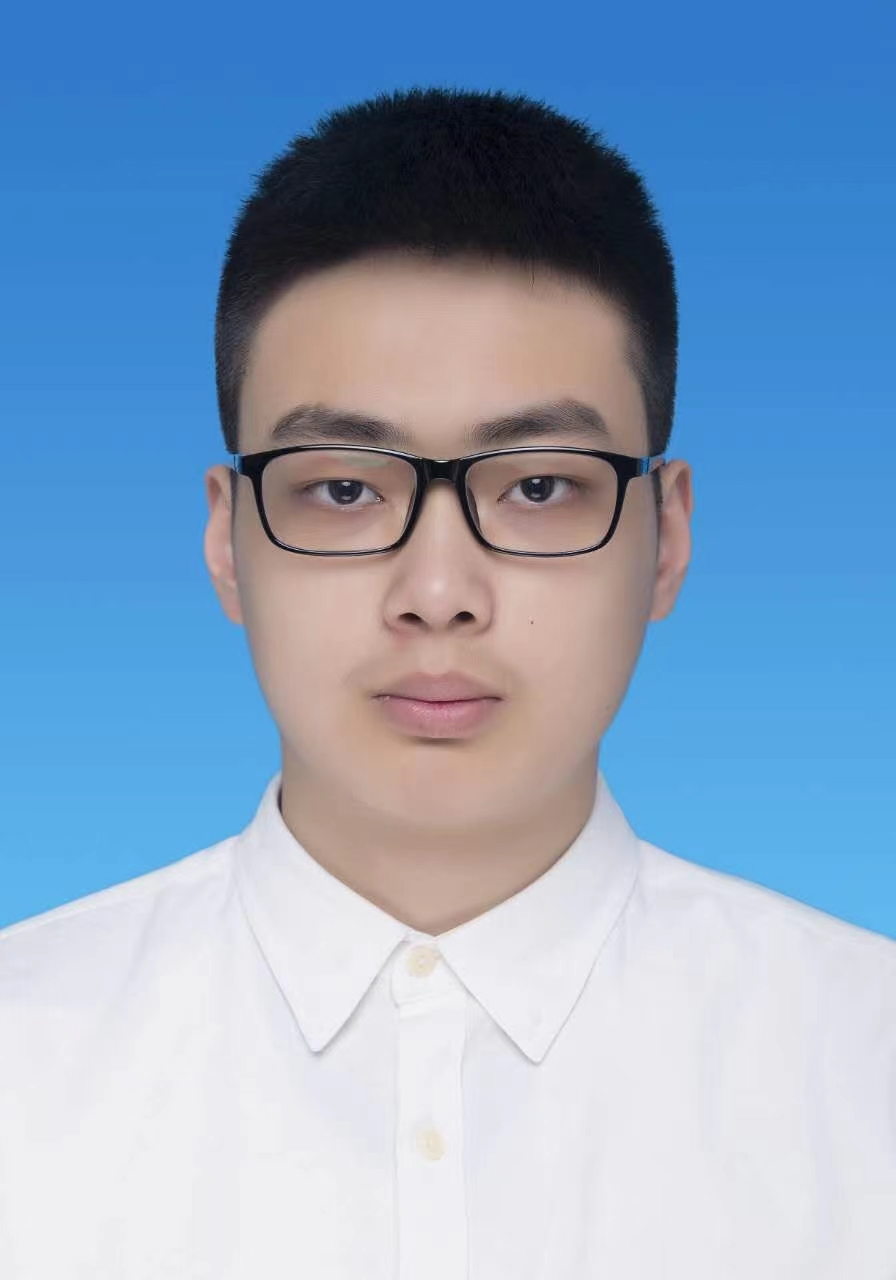}}]{Zherui Lu}
    received the bachelor's degree in detection guidance and control engineering from the Beihang University, Beijing, China, in 2017.
    He is currently pursuing the master's degree with the School of Instrumentation and Optoelectronic Engineering, Beihang University, Beijing.
    His research interests include dim target detection and image processing.
\end{IEEEbiography}
\vspace{-10mm}
\begin{IEEEbiography}[{\includegraphics[width=1in,height=1.25in,clip,keepaspectratio]{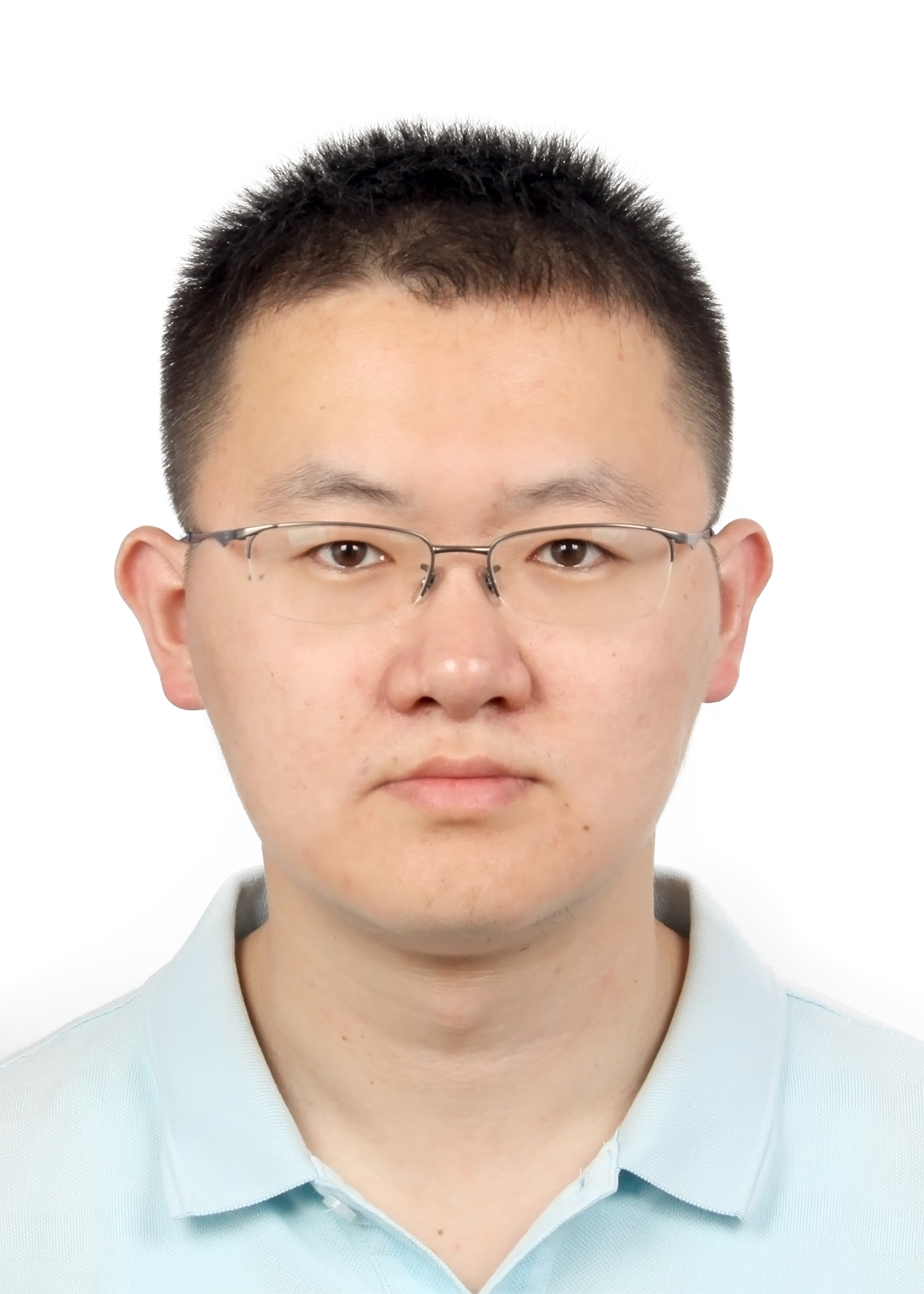}}]{Gangyi Wang}
received the bachelor's degree in electronics and information engineering and the master's degree and Ph.D. degrees in information and communication engineering from the Harbin Institute of Technology. Harbin, China, in 2006, 2008. and 2013, respectively.

He is currently an Associate Professor with the School of Instrumentation and Optoelectronic Engineering, Beihang University, Beijing, China. His research interests include star tracker, computer vision, and embedded systems.
\end{IEEEbiography}
\vspace{-10mm}
\begin{IEEEbiography}[{\includegraphics[width=1in,height=1.25in,clip,keepaspectratio]{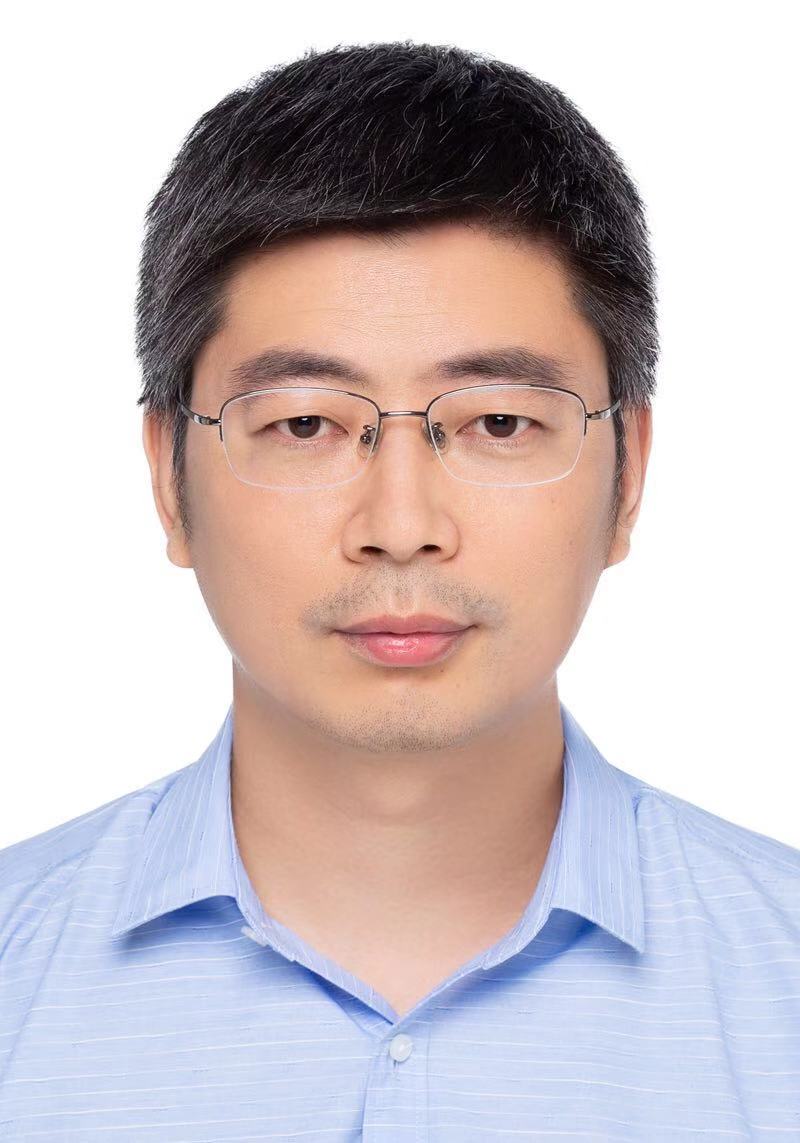}}]{Xinguo Wei}
    was born in 1977. He received the bachelor’s and Ph.D. degrees from the School of Automation Science and Electrical Engineering, Beihang University, Beijing, China, in 1999 and 2004, respectively.

    He was appointed as a Changjiang Distinguished Professor by the Ministry of Education in 2017. He is currently a Professor with the School of Instrumentation and Optoelectronic Engineering, Beihang University. He has authored over 90 articles and over 20 inventions. His research interests include precision measurement, machine vision, and image processing.
\end{IEEEbiography}
\vspace{-10mm}
\begin{IEEEbiography}[{\includegraphics[width=1in,height=1.25in,clip,keepaspectratio]{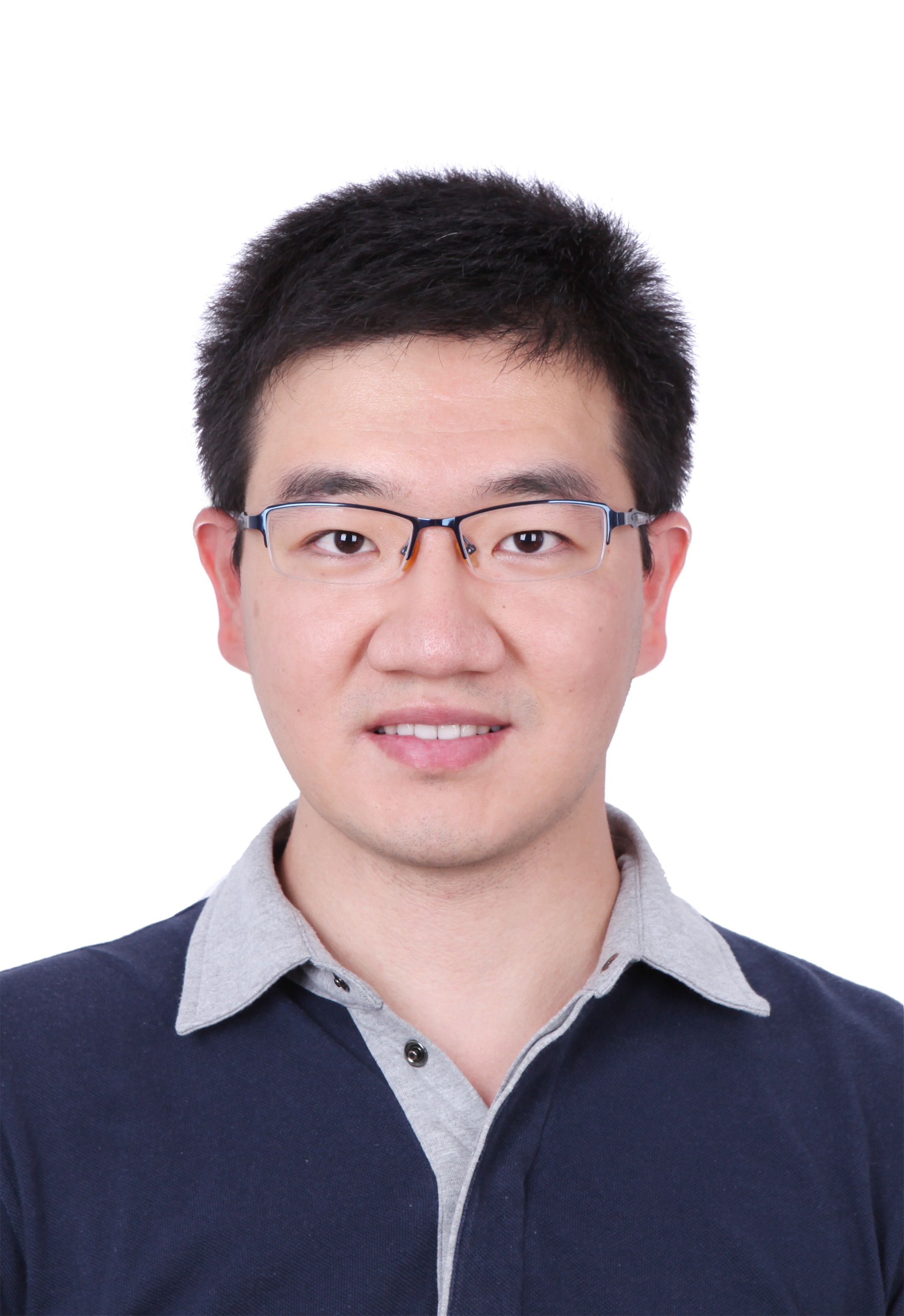}}]{Jian Li}
received the bachelor’s and Ph.D. degrees in precision instrument from Beihang University, Beijing, China, in 2008 and 2015, respectively.
He is currently a Lecturer with the School of Instrumentation and Optoelectronic Engineering, Beihang University. His research interests include
attitude determination sensing systems and image processing.
    
\end{IEEEbiography}

\vspace{-10mm}

\vfill

\end{document}